%% file: main.tex
\documentclass{article} % For LaTeX2e
\usepackage{iclr2027_conference,times}

\usepackage{hyperref}
\usepackage{url}

\usepackage{amsmath}
\usepackage{amssymb}
\usepackage{amsfonts}

\usepackage{booktabs}
\usepackage{multirow}
\usepackage{graphicx}
\usepackage[table]{xcolor}
\usepackage{array}
\usepackage{tabularx}
\usepackage{adjustbox}

\usepackage{pifont}
\usepackage{placeins}
\usepackage{float}
\usepackage[capitalize]{cleveref}
\crefname{section}{Sec.}{Secs.}
\Crefname{section}{Section}{Sections}
\crefname{table}{Tab.}{Tabs.}
\Crefname{table}{Table}{Tables}
\crefname{appendix}{Appendix}{Appendices}
\Crefname{appendix}{Appendix}{Appendices}
\hypersetup{
    pdfauthor={},
    pdftitle={Revisiting Numerical Forecasting Models for Language-Based Human Trajectory Prediction}
}
\newcommand{\pslp}{/}
\newcommand{\tbf}[1]{\textbf{#1}}
\newcommand{\tul}[1]{\underline{#1}}
\newcommand{\NA}{\textcolor{gray}{N/A}}

\newcommand{\MSE}{\operatorname{MSE}}

\newsavebox{\MoRELeftCaption}
\newsavebox{\MoRERightCaption}
\newsavebox{\MoRELeftBody}
\newsavebox{\MoRERightBody}
\newcommand{\MoREPairCheck}[1]{%
  \typeout{MoRE #1 caption heights: \the\dimexpr\ht\MoRELeftCaption+\dp\MoRELeftCaption\relax,
    \the\dimexpr\ht\MoRERightCaption+\dp\MoRERightCaption\relax}%
  \typeout{MoRE #1 body heights: \the\dimexpr\ht\MoRELeftBody+\dp\MoRELeftBody\relax,
    \the\dimexpr\ht\MoRERightBody+\dp\MoRERightBody\relax}%
}

\title{Revisiting Numerical Forecasting Models for Language-Based Trajectory Prediction}
\author{
JunGyu Lee$^1$ \quad Inhwan Bae$^2$ \quad Hae-Gon Jeon$^{1}$\thanks{Corresponding author.} \\
$^1$Yonsei University \quad $^2$DGIST
}
\begin{document}
\iclrfinalcopy
\maketitle
% \lhead{Under review as a conference paper at ICLR 2027}
\lhead{Preprint.}

\begin{abstract}
Language-based trajectory predictors represent coordinates as discrete tokens and learn auxiliary tasks such as destination and group reasoning. This formulation enables the model to capture behavioral intent and social context beyond coordinate dynamics alone. However, token-level objectives provide only indirect guidance for continuous coordinate-space dynamics. To address this limitation, we introduce \textbf{MoRE} (\textbf{M}ixture \textbf{o}f \textbf{R}eward \textbf{E}xperts), a refinement framework that transfers numerical forecasting priors into a pretrained language-based predictor through reinforcement learning. Five frozen numerical predictors provide complementary coordinate-level knowledge of motion and interactions. Their predictions are converted into expert rewards and combined through an uncertainty-weighted consensus that penalizes disagreement. A ground-truth reward anchors the prediction to the target trajectory. To focus refinement on difficult cases, MoRE refines the policy using the top 1\% of training samples ranked by predictive entropy. Expert predictions are computed once and cached before PPO training, so the experts are not run during policy updates or inference. In this way, MoRE combines the contextual modeling of the language-based predictor with coordinate-level feedback from numerical experts. On ETH-UCY, MoRE reduces ADE from 0.22 to 0.20 m and FDE from 0.32 to 0.29 m. Relative to the base policy, ADE  decreases by 17.9\% on SDD and 12.7\% on NBA. On ETH-UCY, MoRE also reduces collision rates and better matches ground-truth pedestrian spacing, without increasing measured inference memory or latency. The project page is available at \url{https://jungyu0413.github.io/MoRE/}.
\end{abstract}

\begin{figure}[ht]
\centering
\includegraphics[width=\linewidth]{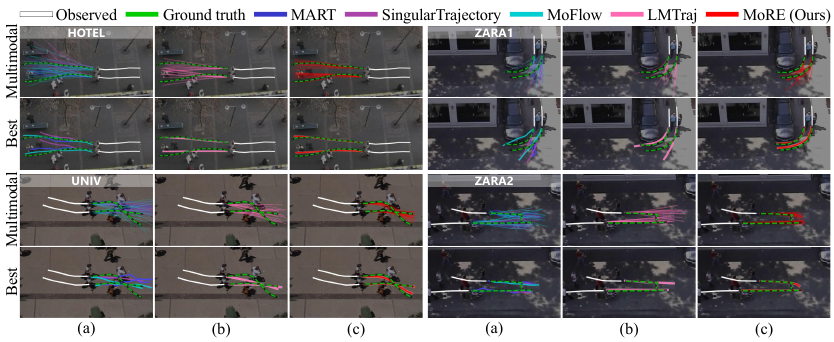}
\vspace{-6mm}
\caption{Prediction examples on ETH-UCY. ``Multimodal'' shows 20 samples, and ``Best'' shows the closest sample. (a) Numerical baselines and (b) LMTraj exhibit different spatial patterns. (c) MoRE refines the language-based predictions using numerical feedback.}
\label{fig:motivation}
\end{figure}

\section{Introduction}\label{sec:intro}
Forecasting human trajectories requires modeling motion, interactions, and uncertainty about future behavior. Numerical trajectory predictors model human motion directly in continuous coordinate space and have progressively incorporated specialized priors for local interactions, group behavior, motion patterns, and destination goals~\citep{mohamed2020social,bae2022gpgraph,bae2024singulartrajectory,zhao2021experttraj}. This coordinate-based formulation provides strong spatial accuracy, particularly for the best sample under standard best-of-$K$ evaluation (\cref{fig:motivation}(a), Best). However, coordinate-based numerical predictors do not directly leverage the contextual knowledge captured by language-based models. As a result, stochastic numerical predictors often generate high-variance, scattered predictive distributions while optimizing coordinate-level regression objectives (\cref{fig:motivation}(a), Multimodal).

Language-based predictors provide a complementary formulation by representing trajectories and related reasoning tasks as discrete tokens. LMTraj~\citep{bae2024lmtrajectory} jointly learns destination, group, collision, and other auxiliary tasks, allowing the model to capture behavioral intent and social context more explicitly. Compared with the dispersed multimodal predictions of the numerical baselines in \cref{fig:motivation}(a), LMTraj produces more spatially consistent multimodal paths in \cref{fig:motivation}(b). However, token-level likelihood does not directly reflect metric errors in continuous coordinates, so its best prediction can remain spatially inaccurate (\cref{fig:motivation}(b), Best). This motivates incorporating coordinate-level numerical feedback during refinement. Our goal is to improve the coordinate accuracy of a pretrained language-based predictor using numerical forecasting knowledge, without requiring numerical experts at inference.

We propose MoRE (Mixture of Reward Experts), a framework that transfers complementary knowledge from pretrained numerical predictors to a language-based predictor during training. Rather than directly combining expert outputs, MoRE converts their predictions into reward signals, integrating the language-based model's contextual knowledge with the complementary motion and social-interaction priors learned by numerical predictors. The expert rewards are combined through a consensus objective and used to refine the policy with reinforcement learning. To keep the process efficient, refinement focuses on difficult samples identified by predictive uncertainty, with expert predictions computed and cached in advance.

MoRE combines the contextual representations of the language-based predictor with complementary knowledge from multiple numerical experts. This integration improves prediction accuracy across ETH-UCY, SDD, GCS, and NBA, with larger gains on challenging non-linear motion and cross-dataset transfer. Overall, MoRE transfers knowledge from multiple numerical experts into a single language-based predictor during training, without requiring the experts at inference or increasing the measured inference memory and latency of the base policy.

\section{Related Work}\label{sec:related}

\subsection{Numerical Trajectory Prediction}

Numerical trajectory prediction has progressed from hand-crafted social interaction models~\citep{helbing1995social,pellegrini2009you} to data-driven approaches that learn interactions and multimodal futures~\citep{alahi2016social,gupta2018social,mangalam2020pecnet,salzmann2020trajectron++}. Recent methods introduce different inductive biases to capture human motion more effectively. Spatio-temporal graph models focus on interaction structure~\citep{mohamed2020social,huang2019stgat,bae2021dmrgcn}, while GP-Graph and GroupNet explicitly model group-level relations~\citep{bae2022gpgraph,xu2022groupnet}. Other methods capture characteristic motion patterns through compact trajectory representations~\citep{bae2023eigentrajectory,bae2024singulartrajectory}, or guide future motion using destination information~\citep{dendorfer2020goalgan,zhao2021experttraj}. Recent generative and relational approaches further improve multimodal forecasting through diffusion, transformer, and flow-based modeling~\citep{gu2022mid,mao2023leapfrog,lee2024mart,fu2025moflow}. These numerical predictors already model social interactions, group structure, and diverse motion patterns~\citep{kothari2021human}, but they encode such knowledge through different architectures and objectives. MoRE leverages this diversity rather than introducing another numerical predictor. We use models with complementary strengths as numerical experts and transfer their specialized coordinate-level knowledge to a language-based predictor.

\subsection{Language-Based Trajectory Prediction}

Language-based trajectory prediction represents coordinates as tokens and formulates forecasting as sequence generation. LMTraj~\citep{bae2024lmtrajectory} jointly learns trajectory prediction with auxiliary tasks involving destinations, directions, groups, collisions, and mimicry, while its multimodal extension incorporates visual information~\citep{bae2025vlmtrajectory}. Subsequent approaches explore language-guided numerical forecasting~\citep{chib2025lgtraj}, goal-oriented prompting and chain-of-thought reasoning~\citep{kim2025guidecot}, visual-linguistic reasoning~\citep{shenkut2025visual}, autoregressive multimodal generation~\citep{wang2026autotraces}, and reinforcement learning for language-based trajectory prediction~\citep{xu2026w2w}.
These methods aim to strengthen contextual and behavioral reasoning, but token-only objectives provide only indirect supervision for continuous coordinate accuracy. MoRE combines the contextual modeling of the language-based predictor with coordinate-level feedback from numerical predictors while retaining a single language-based predictor at inference.

\subsection{Expert Transfer and Reward-Based Refinement}
Knowledge from multiple models can be transferred in several ways. Knowledge distillation trains a student to reproduce the predictions of one or more teacher models~\citep{hinton2015distilling,you2017multiple}. Mixture-of-experts methods instead maintain subnetworks and select or combine them within the model~\citep{jacobs1991adaptive,shazeer2017moe,fedus2022switch}. These approaches transfer expert knowledge through supervised targets or architectural composition. Reward-based learning provides another form of supervision by evaluating generated outputs and optimizing the policy from this feedback~\citep{christiano2017rlhf,ouyang2022instructgpt}. Since a reward can be an imperfect proxy for the desired behavior, aggressive optimization may lead to reward overoptimization~\citep{gao2023scaling}. Reward-model ensembles reduce reliance on a single reward signal, and \citet{coste2024reward} investigate conservative aggregation strategies including uncertainty-weighted optimization. MoRE builds on this reward-based approach but uses pretrained numerical predictors as domain experts. Their forecasts provide coordinate-level rewards for refining a language-based predictor, rather than serving as direct supervision targets or inference-time modules.

\section{Method}\label{sec:method}

MoRE builds on a pretrained language-based forecasting architecture and refines it using rewards from numerical experts. We first describe the tokenized trajectory prediction setup (\cref{sec:prelim}), followed by the multi-expert consensus reward (\cref{sec:consensus}), PPO-based refinement (\cref{sec:training}), and uncertainty-driven sample selection (\cref{sec:mining}). Implementation details are provided in \cref{sec:implementation}.

\begin{figure}[!htbp]
\centering
\includegraphics[width=\linewidth]{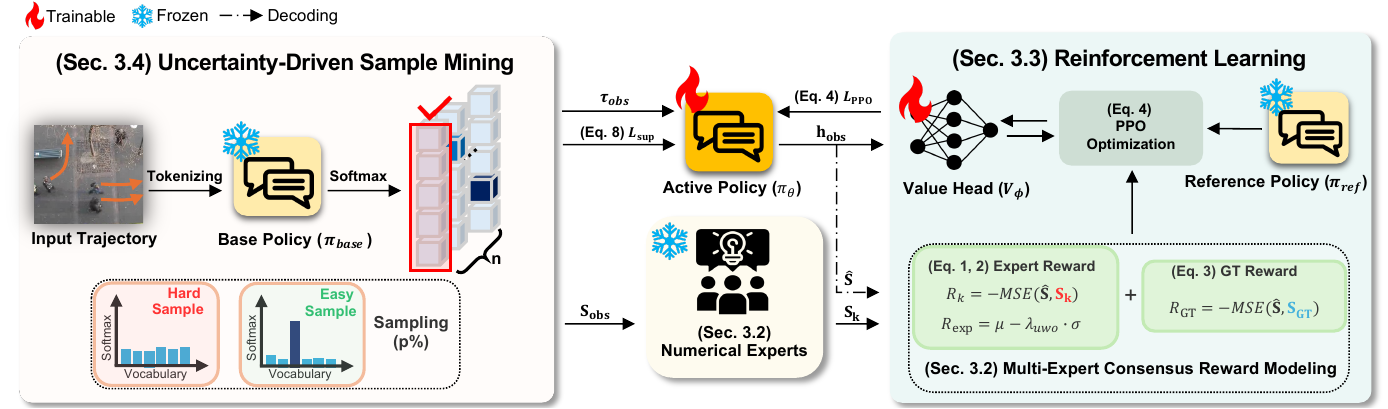}
\vspace{-5mm}
\caption{MoRE training pipeline. A frozen base policy ranks training samples by predictive entropy. Numerical expert predictions for the selected subset are cached once. Decoded policy samples receive expert-consensus and ground-truth rewards for PPO refinement, interleaved with supervised learning. Only the refined forecasting policy is needed at inference.}
\label{fig:framework}
\vspace{-4mm}
\end{figure}

\subsection{Preliminaries}\vspace{-1mm}
\label{sec:prelim}
\vspace{0mm}\noindent\textbf{Problem Definition.} 
Given observed trajectories of $N$ pedestrians over $T_{\text{obs}}$ timesteps, our goal is to predict their future trajectories over $T_{\text{pred}}$ timesteps. Formally, let $\mathbf{p}_t^n = [x_t^n, y_t^n]^\top \in \mathbb{R}^2$ denote the 2D coordinate vector of pedestrian $n$ at time $t$. The observed trajectory and the ground truth future trajectory for pedestrian $n$ are defined as matrices $\mathbf{S}^n_{\text{obs}} = [\mathbf{p}_1^n, \dots, \mathbf{p}_{T_{\text{obs}}}^n]^\top \in \mathbb{R}^{T_{\text{obs}} \times 2}$ and $\mathbf{S}^n_{\text{GT}} = [\mathbf{p}_{T_{\text{obs}}+1}^n, \dots, \mathbf{p}_{T_{\text{obs}}+T_{\text{pred}}}^n]^\top \in \mathbb{R}^{T_{\text{pred}} \times 2}$, respectively. Consequently, our model aims to learn the parameterized conditional distribution across all $N$ pedestrians, formulated as $p_\theta(\{ \mathbf{S}^n_{\text{GT}} \}_{n=1}^N \mid \{ \mathbf{S}^n_{\text{obs}} \}_{n=1}^N)$. For the pedestrian benchmarks, we use $T_{\text{obs}}=8$ and $T_{\text{pred}}=12$. NBA follows its benchmark-specific evaluation protocol described in \cref{sec:setup}.

\vspace{2mm}\noindent\textbf{Base Policy.}
Our framework builds upon LMTraj~\citep{bae2024lmtrajectory}, which formulates trajectory prediction as a text-based sequence generation task~\citep{raffel2020t5}. A numerical tokenizer maps continuous trajectories into a discrete token space, and we initialize from its pretrained weights. However, next-token likelihood does not directly penalize the magnitude of errors in decoded coordinates. Formally, the tokenizer maps the continuous observation matrix $\mathbf{S}_{\text{obs}}$ to a discrete text prompt $\tau_{\text{obs}}$, and the ground-truth future trajectory $\mathbf{S}_{\text{GT}}$ to a target token sequence $\tau_{\text{GT}}=[\tau_1,\dots,\tau_L]$. The policy models the discrete conditional distribution $\pi_\theta(\tau_k\mid\tau_{\text{obs}},\tau_{<k})$.

\subsection{Multi-Expert Consensus Reward Modeling}
\label{sec:consensus}

\vspace{0mm}\noindent\textbf{Numerical Reward Formulation.}
We define a composite reward that integrates complementary knowledge from multiple numerical experts. We use five specialized predictors with different strengths: 
(i) a \textit{local interaction expert} for local pedestrian interactions,
(ii) a \textit{structural relation expert} for relational dynamics,
(iii) a \textit{social grouping expert} for group behavior,
(iv) a \textit{motion pattern expert} for characteristic motion patterns, and
(v) a \textit{goal expert} for destination information.
Each expert contributes its own prediction as coordinate-level guidance, allowing MoRE to preserve their complementary knowledge rather than forcing them to agree on a single behavior. At the same time, individual predictions can contain noisy or unreliable guidance. MoRE therefore applies a consensus reward in which larger disagreement among expert scores receives a stronger penalty, reducing the influence of strongly inconsistent signals while retaining the guidance from each expert. The same framework can incorporate other pretrained numerical predictors, and results with additional experts and alternative expert configurations are provided in \cref{sec:supp_additional_experiments}.

\vspace{2mm}\noindent\textbf{Reward Aggregation via Uncertainty-Weighted Optimization.}
Given a text-based observation prompt, the language-based policy autoregressively generates a sequence of discrete tokens, which is decoded into a continuous coordinate matrix $\hat{\mathbf{S}}$. For each selected observation, each frozen numerical expert $k \in \{1,\ldots,K\}$ produces a continuous prediction $\mathbf{S}_k$ from the historical trajectory, and these predictions are cached before policy refinement. We then define the reward from expert $k$ as the negative mean squared error between the decoded policy prediction and the cached expert prediction:
\begin{equation}
    R_k = -\text{MSE}(\hat{\mathbf{S}}, \mathbf{S}_k).
\end{equation}

To combine expert rewards, we follow the conservative aggregation principle of Uncertainty-Weighted Optimization (UWO)~\citep{coste2024reward}. For each generated trajectory, the mean reward represents its overall agreement with the numerical experts, while the standard deviation measures disagreement among their scores. We penalize this disagreement so that trajectories receiving inconsistent evaluations from the experts obtain lower consensus rewards. The resulting consensus reward is defined as:
\begin{equation}
    R_{\text{exp}} = \mu - \lambda_{\text{uwo}} \cdot \sigma, \quad \text{where } \mu = \frac{1}{K} \sum_{k=1}^{K} R_k, \;\; \sigma = \sqrt{\frac{1}{K} \sum_{k=1}^{K} (R_k - \mu)^2}.
\end{equation}
The penalty coefficient $\lambda_{\text{uwo}}$ reduces the influence of noisy or inconsistent expert signals by penalizing disagreement among expert rewards, encouraging more stable and physically grounded predictions. Here, $\mu$ is the mean expert reward and $\sigma$ is the standard deviation of expert rewards for the same generated trajectory. Setting $\lambda_{\text{uwo}}=0$ recovers mean aggregation.

\vspace{2mm}\noindent\textbf{Total Reward Formulation.}
To balance expert guidance with accuracy to the ground-truth trajectory, we use the composite reward:
\begin{equation}
\label{eq:total_reward}
    R = R_{\text{GT}} + \lambda_{\text{exp}} \cdot R_{\text{exp}},
    \quad
    \text{where } R_{\text{GT}}
    = -\text{MSE}(\hat{\mathbf{S}}, \mathbf{S}_{\text{GT}}).
\end{equation}
While $R_{\text{GT}}$ anchors the prediction to the target trajectory, $R_{\text{exp}}$ incorporates complementary knowledge from the numerical experts. The expert reward encourages agreement with their forecasts rather than directly enforcing physical constraints.

\subsection{Reinforcement Learning via PPO}\label{sec:training}

\vspace{0mm}\noindent\textbf{PPO Optimization.}
The policy $\pi_\theta$ and the value network $V_\phi$ are jointly optimized by minimizing a composite surrogate loss, which integrates a clipped policy objective, a value regression loss, and a KL divergence penalty to prevent catastrophic forgetting:
\begin{equation}
\label{eq:ppo}
\begin{aligned}
    \mathcal{L}_{\text{PPO}} =& -\mathbb{E}\left[ \frac{1}{L} \sum_{t=1}^{L} \min\left(\rho_t(\theta) \hat{A}, \text{clip}(\rho_t(\theta), 1-\epsilon, 1+\epsilon) \hat{A}\right) \right] + c_v \mathbb{E}\left[ (V_\phi(\mathbf{h}_{\text{obs}}) - R)^2 \right] \\
    & + \beta \hat{D}_{\text{KL}}(\pi_\theta \| \pi_{\text{ref}}), \quad \text{where} \quad \rho_t(\theta) = \frac{\pi_\theta(\tau_t \mid \tau_{\text{obs}}, \tau_{<t})}{\pi_{\text{old}}(\tau_t \mid \tau_{\text{obs}}, \tau_{<t})}.
\end{aligned}
\end{equation}
The token-level importance ratio $\rho_t(\theta)$ is computed for each generation step $t$ relative to the behavior policy $\pi_{\text{old}}$. The sequence-level advantage $\hat{A} = R - V_\phi(\mathbf{h}_{\text{obs}})$ is calculated as the direct difference between the composite reward $R$ and the expected return. The value function $V_\phi$ processes the mean-pooled encoder hidden states of the observation context ($\mathbf{h}_{\text{obs}}$) and is jointly trained via Mean Squared Error (MSE) regression on the observed returns $R$. Further mathematical details are provided in \cref{sec:supp_rl_details}.

\subsection{Uncertainty-Driven Sample Mining}
\label{sec:mining}
We focus PPO refinement on training samples with high predictive entropy under the frozen base policy. Entropy is used to rank sample uncertainty, and numerical expert rewards are applied to the selected top-$p$\% subset.

\vspace{2mm}\noindent\textbf{Sample Difficulty Scoring.} 
We assess the difficulty of each trajectory sample by measuring the predictive uncertainty of the base policy $\pi_{\text{base}}$. Rather than relying on ground-truth predictive error, we compute the Shannon entropy of the model's Softmax output distribution to reveal inherent model ambiguity. The difficulty score $\mathcal{S}_i$ for sample $i$ is defined as the sequence-averaged entropy across the vocabulary space $\mathcal{V}$:
\begin{equation}
    \mathcal{S}_i = -\frac{1}{L_i} \sum_{k=1}^{L_i} \sum_{v \in \mathcal{V}} \pi_{\text{base}}(v \mid \tau_{i,\text{obs}}, \tau_{i,<k}) \log \pi_{\text{base}}(v \mid \tau_{i,\text{obs}}, \tau_{i,<k}),
    \label{eq:entropy}
\end{equation}

where $v$ represents a candidate token and $L_i$ is the sequence length. Here, a high $\mathcal{S}_i$ indicates a relatively flat Softmax distribution, indicating greater uncertainty in the base policy's token distribution. Based on this difficulty distribution across the entire training set $\mathcal{D}$, we define a selection threshold $\gamma_p$ to identify the top $p\%$ of samples. We then apply a binary selection mask $M_i$ to focus the optimization processes on these high-uncertainty cases:

\begin{equation}
    \gamma_p = \text{Percentile}(\{\mathcal{S}_i\}_{i=1}^{|\mathcal{D}|}, 100-p), \quad M_i = \mathbb{1}[\mathcal{S}_i \geq \gamma_p].
    \label{eq:mining}
\end{equation}

\vspace{2mm}\noindent\textbf{Overall Objective.}
The refinement process combines supervised learning with PPO on the selected subset. The total objective is defined as:
\begin{equation}
    \mathcal{L}_{\text{total}} = \frac{1}{|M|} \sum_{i \in M} \left( \mathcal{L}_{\text{sup}}^{(i)} + \lambda_{\text{RL}} \mathcal{L}_{\text{PPO}}^{(i)} \right),
    \label{eq:total_loss}
\end{equation}
where $M = \{i : M_i = 1\}$ denotes the identified subset of hard samples. The supervised term $\mathcal{L}_{\text{sup}}^{(i)}$ acts as a regularizer to maintain the model's fundamental sequence-to-sequence mapping capabilities. It is defined as the standard negative log-likelihood (NLL) against the ground-truth token sequence $\tau_{i}^*$:
\begin{equation}
    \mathcal{L}_{\text{sup}}^{(i)} = - \frac{1}{L_i} \sum_{k=1}^{L_i} \log \pi_\theta(\tau_{i,k}^* \mid \tau_{i,\text{obs}}, \tau_{i,<k}^*).
\end{equation}
The PPO term provides reward-based feedback from the ground-truth and numerical expert rewards. During training, PPO and supervised updates are interleaved, with one PPO update every ten supervised steps. This schedule maintains the original token-prediction objective while introducing coordinate-level guidance through PPO.

\subsection{Implementation Details}
\label{sec:implementation}
We adopt the pretrained LMTraj~\citep{bae2024lmtrajectory} as the base policy and refine it with LoRA~\citep{hu2022lora} of rank $r=16$, updating 0.3\% of the base-policy parameters. The five frozen numerical experts are Social-STGCNN~\citep{mohamed2020social}, DMRGCN~\citep{bae2021dmrgcn}, GP-Graph~\citep{bae2022gpgraph}, SingularTrajectory~\citep{bae2024singulartrajectory}, and Expert-Trajectory~\citep{zhao2021experttraj}. We set $\lambda_{\text{exp}}=0.5$, $\lambda_{\text{uwo}}=0.05$, and $\lambda_{\text{RL}}=0.5$. PPO uses $\epsilon=0.2$, $c_v=0.5$, and $\beta=0.1$. We refine the top 1\% of training samples ranked by predictive entropy using AdamW~\citep{loshchilov2017decoupled} with a learning rate of $10^{-4}$ and an effective batch size of 1,024. Refinement takes about one hour on eight NVIDIA RTX PRO 6000 GPUs.

\begin{table}[!b]
\vspace{-4mm}
\centering
\caption{Best-of-20 ADE / FDE (ETH-UCY in meters, with SDD and GCS in pixels). Bold and underline indicate best and second-best values, including ties. Model references are in \cref{sec:sota_sources}.}
\label{tab:sota_comparison}
\smallskip

\footnotesize
\setlength{\tabcolsep}{2pt}%

\begin{tabular*}{\linewidth}{
@{}l@{\extracolsep{\fill}}
cccccc
!{\vrule width 0.3pt}
c
!{\vrule width 0.3pt}
c@{}
}
\toprule
\textbf{Model}
& \textbf{ETH}
& \textbf{HOTEL}
& \textbf{UNIV}
& \textbf{ZARA1}
& \textbf{ZARA2}
& \textbf{AVG}
& \textbf{SDD}
& \textbf{GCS} \\
\midrule

\rowcolor{gray!10}
\multicolumn{9}{l}{\textit{Numerical models}} \\

Social GAN
& 0.77\pslp 1.40
& 0.43\pslp 0.88
& 0.75\pslp 1.50
& 0.35\pslp 0.69
& 0.36\pslp 0.72
& 0.53\pslp 1.04
& 13.6\pslp 24.6
& 15.9\pslp 32.6 \\

Social-STGCNN
& 0.65\pslp 1.10
& 0.50\pslp 0.86
& 0.44\pslp 0.80
& 0.34\pslp 0.53
& 0.31\pslp 0.48
& 0.45\pslp 0.75
& 20.8\pslp 33.2
& 14.7\pslp 23.9 \\

PECNet
& 0.61\pslp 1.07
& 0.22\pslp 0.39
& 0.34\pslp 0.56
& 0.25\pslp 0.45
& 0.19\pslp 0.33
& 0.32\pslp 0.56
& 10.0\pslp 15.9
& 17.1\pslp 29.3 \\

Trajectron++
& 0.61\pslp 1.03
& 0.20\pslp 0.28
& 0.30\pslp 0.55
& 0.24\pslp 0.41
& 0.18\pslp 0.32
& 0.31\pslp 0.52
& 11.4\pslp 20.1
& 12.8\pslp 24.2 \\

AgentFormer
& 0.46\pslp 0.80
& 0.14\pslp 0.22
& 0.25\pslp 0.45
& 0.18\pslp 0.30
& 0.14\pslp 0.24
& 0.23\pslp 0.40
& 8.7\pslp 14.9
& 10.2\pslp 16.9 \\

MID
& 0.57\pslp 0.93
& 0.21\pslp 0.33
& 0.29\pslp 0.55
& 0.28\pslp 0.50
& 0.20\pslp 0.37
& 0.31\pslp 0.54
& 7.6\pslp 14.3
& 10.7\pslp 18.2 \\

EqMotion
& 0.40\pslp 0.61
& \tul{0.12}\pslp 0.18
& 0.23\pslp 0.43
& 0.18\pslp 0.32
& \tul{0.13}\pslp 0.23
& \tul{0.21}\pslp 0.35
& 7.9\pslp 11.9
& 7.6\pslp 13.1 \\

GP-Graph
& 0.43\pslp 0.63
& 0.18\pslp 0.30
& 0.24\pslp 0.42
& \tul{0.17}\pslp 0.31
& 0.15\pslp 0.29
& 0.23\pslp 0.39
& 9.1\pslp 13.8
& 7.8\pslp 13.7 \\

LED
& 0.39\pslp 0.58
& \tbf{0.11}\pslp 0.17
& 0.26\pslp 0.43
& 0.18\pslp \tbf{0.26}
& \tul{0.13}\pslp \tul{0.22}
& \tul{0.21}\pslp 0.33
& 8.5\pslp 11.7
& 9.0\pslp 11.4 \\

MART
& \tbf{0.35}\pslp 0.47
& 0.14\pslp 0.22
& 0.25\pslp 0.45
& \tul{0.17}\pslp 0.29
& \tul{0.13}\pslp \tul{0.22}
& \tul{0.21}\pslp 0.33
& \tul{7.4}\pslp 11.8
& 10.6\pslp 14.1 \\

SingularTraj.
& \tbf{0.35}\pslp \tul{0.42}
& 0.13\pslp 0.19
& 0.25\pslp 0.44
& 0.19\pslp 0.32
& 0.15\pslp 0.25
& \tul{0.21}\pslp 0.32
& 7.58\pslp 12.1
& 7.9\pslp 13.3 \\

MoFlow
& 0.40\pslp 0.57
& \tbf{0.11}\pslp 0.17
& 0.23\pslp 0.39
& \tbf{0.15}\pslp \tbf{0.26}
& \tbf{0.12}\pslp \tul{0.22}
& \tbf{0.20}\pslp 0.32
& 7.5\pslp 12.0
& 9.1\pslp 11.6 \\

PCHGCN
& 0.42\pslp 0.65
& 0.17\pslp 0.28
& \tul{0.21}\pslp 0.38
& \tul{0.17}\pslp 0.31
& \tul{0.13}\pslp 0.23
& 0.22\pslp 0.37
& \NA
& \NA \\

ViTE
& \tbf{0.35}\pslp 0.49
& \tbf{0.11}\pslp 0.17
& 0.23\pslp 0.42
& 0.18\pslp 0.30
& \tul{0.13}\pslp \tul{0.22}
& \tbf{0.20}\pslp 0.32
& \tul{7.4}\pslp 11.9
& \NA \\

\midrule

\rowcolor{gray!10}
\multicolumn{9}{l}{\textit{Language-based models}} \\

LMTraj-ZERO
& 0.80\pslp 1.64
& 0.20\pslp 0.37
& 0.37\pslp 0.77
& 0.33\pslp 0.66
& 0.24\pslp 0.50
& 0.39\pslp 0.79
& 10.9\pslp 21.0
& 12.7\pslp 25.5 \\

LMTraj-SUP
& 0.41\pslp 0.50
& \tul{0.12}\pslp 0.16
& 0.22\pslp \tul{0.34}
& 0.20\pslp 0.32
& 0.17\pslp 0.27
& 0.22\pslp 0.32
& 7.8\pslp \tul{10.1}
& \tul{7.1}\pslp \tul{9.6} \\

GUIDE-CoT
& 0.38\pslp 0.43
& 0.13\pslp \tul{0.15}
& 0.34\pslp 0.48
& 0.19\pslp 0.29
& 0.17\pslp \tbf{0.21}
& 0.24\pslp \tul{0.31}
& \NA
& \NA \\

W2W
& \tbf{0.35}\pslp \tbf{0.41}
& \tul{0.12}\pslp \tul{0.15}
& \tbf{0.20}\pslp \tbf{0.32}
& 0.19\pslp 0.29
& 0.17\pslp 0.26
& \tul{0.21}\pslp \tbf{0.29}
& \tul{7.4}\pslp \tul{10.1}
& \NA \\

\midrule

\textbf{MoRE (Ours)}
& \tul{0.36}\pslp \tul{0.42}
& \tbf{0.11}\pslp \tbf{0.14}
& \tul{0.21}\pslp \tbf{0.32}
& 0.18\pslp \tul{0.28}
& 0.17\pslp 0.26
& \tbf{0.20}\pslp \tbf{0.29}
& \tbf{6.4}\pslp \tbf{9.5}
& \tbf{6.8}\pslp \tbf{8.5} \\

\bottomrule
\end{tabular*}

\vspace{-2mm}
\end{table}

\section{Experiments}\label{sec:experiments}
\subsection{Experimental Setup}\label{sec:setup}

\noindent\textbf{Benchmarks and protocol.}
We evaluate on ETH-UCY~\citep{pellegrini2009you,lerner2007crowdsbyexample}, Stanford Drone Dataset (SDD)~\citep{robicquet2016learning}, Grand Central Station (GCS)~\citep{yi2015understanding}, and NBA. ETH-UCY comprises ETH, HOTEL, UNIV, ZARA1, and ZARA2 and uses leave-one-out evaluation. SDD contains aerial trajectories, while GCS includes dense station scenes with up to 289 concurrent pedestrians. These pedestrian benchmarks use eight observed and twelve future frames. NBA evaluates highly interactive motion under the protocol of \citet{wong2025resonance}. Cross-dataset evaluation trains on ETH-UCY and tests on SDD without fine-tuning, assessing whether numerical refinement improves spatial accuracy while retaining the cross-dataset transfer capability of the language-based predictor.

\noindent\textbf{Metrics.}
Average Displacement Error (ADE) measures mean Euclidean position error over future timesteps, while Final Displacement Error (FDE) measures endpoint error. Following the standard best-of-20 evaluation protocol for stochastic trajectory prediction~\citep{gupta2018social}, we report the minimum ADE and FDE over 20 sampled trajectories.

\noindent\textbf{Social spacing and collisions.}
On ETH-UCY, we additionally report collision rate (COL)~\citep{liu2021snce} and pedestrian-spacing statistics based on Hall's proxemic zones~\citep{hall1969hidden}. These distance ranges provide a simple way to compare whether predicted pedestrian spacing resembles real interactions beyond displacement error alone. We use three zones: intimate ($<0.45$\,m), personal (0.45--1.2\,m), and social (1.2--3.7\,m). For each zone, we measure the percentage of pedestrians with at least one other pedestrian within that range. A pedestrian may have neighbors in multiple zones, so the percentages do not need to sum to 100\%. Values closer to the ground truth indicate more realistic spacing, while lower COL is better.

\subsection{Evaluation Results}\label{sec:evaluation}

\noindent\textbf{Forecasting accuracy.}
~\cref{tab:sota_comparison} compares numerical and language-based predictors. At the reported precision, MoRE achieves tied-best average ADE and FDE on ETH-UCY and the lowest ADE and FDE on both SDD and GCS among the compared models. Relative to LMTraj-SUP, ETH-UCY ADE decreases from 0.22 to 0.20\,m and FDE from 0.32 to 0.29\,m, corresponding to approximately 9.1\% and 9.4\% reductions. SDD ADE decreases from 7.8 to 6.4 pixels by approximately 17.9\%, while GCS FDE decreases from 9.6 to 8.5 pixels by approximately 11.5\%. Together, these results show consistent improvements across benchmarks, with MoRE matching the best average accuracy on ETH-UCY and yielding larger relative gains on SDD and GCS.

\ifdefined\MoRESocialBox\else\newsavebox{\MoRESocialBox}\fi
\ifdefined\MoRENBABox\else\newsavebox{\MoRENBABox}\fi
\ifdefined\MoRENBAExtra\else\newlength{\MoRENBAExtra}\fi

\begin{table}[h]
\centering
\vspace{-2mm}

\begin{minipage}[t]{0.625\linewidth}\vspace{0pt}
\caption{Social interaction statistics on ETH-UCY (\%). Intimate, personal, and social values are compared with the ground truth, while lower collision rate (COL) is better.}
\label{tab:combined}
\end{minipage}\hfill
\begin{minipage}[t]{0.345\linewidth}\vspace{0pt}
\caption{Trajectory prediction results on the NBA dataset using best-of-20 ADE / FDE (meters).}
\label{tab:nba_main}
\vspace{-2mm}
\end{minipage}

\par\smallskip

\renewcommand{\arraystretch}{1}
\setlength{\tabcolsep}{3pt}

% ---------------------------------------------------------
% Left table: Social interaction statistics
% ---------------------------------------------------------
\sbox{\MoRESocialBox}{%
\begin{tabularx}{0.625\linewidth}{
    @{}
    l
    >{\centering\arraybackslash}X
    >{\centering\arraybackslash}X
    >{\centering\arraybackslash}X
    >{\centering\arraybackslash}X
    @{}
}
\toprule
\textbf{Method}
& \textbf{Intimate}
& \textbf{Personal}
& \textbf{Social}
& \textbf{COL}$\downarrow$ \\
\midrule
Ground truth
& 22.48
& 62.61
& 12.93
& \textcolor{gray}{N/A} \\
\midrule
GP-Graph
& 26.02
& 58.91
& 13.82
& 2.44 \\

LED
& 27.06
& 58.81
& 13.72
& 2.62 \\

SingularTraj.
& 27.10
& 57.25
& 12.12
& 2.96 \\

MART
& 26.71
& 58.14
& 12.14
& \tul{2.23} \\

LMTraj-SUP
& \tul{25.72}
& \tul{59.79}
& \tul{12.58}
& 2.54 \\
\midrule

\textbf{MoRE (Ours)}
& \tbf{23.41}
& \tbf{60.49}
& \tbf{13.08}
& \tbf{2.11} \\
\bottomrule
\end{tabularx}%
}

\def\MoRENBARowStrut{%
  \rule[
    -\dimexpr\dp\strutbox+\MoRENBAExtra/2\relax
  ]{0pt}{
    \dimexpr\ht\strutbox+\dp\strutbox+\MoRENBAExtra\relax
  }%
}

\def\MoRENBABody{%
\begin{tabularx}{0.345\linewidth}{
    @{}
    l
    >{\centering\arraybackslash}X
    >{\centering\arraybackslash}X
    @{}
}
\toprule
\textbf{Method}
& \textbf{ADE}
& \textbf{FDE} \\
\midrule

\MoRENBARowStrut MemoNet
& 1.25
& 1.47 \\

\MoRENBARowStrut E-V2-Net
& 1.26
& 1.64 \\

\MoRENBARowStrut SocialCircle+
& 1.14
& 1.37 \\

\MoRENBARowStrut Resonance
& 1.12
& 1.38 \\

\MoRENBARowStrut LMTraj-SUP
& \tul{1.10}
& \tul{1.21} \\
\midrule

\textbf{MoRE (Ours)}
& \tbf{0.96}
& \tbf{1.11} \\
\bottomrule
\end{tabularx}%
}

\sbox{\MoRENBABox}{\MoRENBABody}

\setlength{\MoRENBAExtra}{\dimexpr
  (
    \ht\MoRESocialBox+\dp\MoRESocialBox
    -\ht\MoRENBABox-\dp\MoRENBABox
  )/5
\relax}

\ifdim\MoRENBAExtra<0pt
  \setlength{\MoRENBAExtra}{0pt}
\fi

\sbox{\MoRENBABox}{\MoRENBABody}

\noindent
\usebox{\MoRESocialBox}
\hfill
\usebox{\MoRENBABox}

\vspace{-1mm}
\end{table}

\begin{figure}[h]
\centering

\includegraphics[trim={0 505mm 0 0},clip, width=\textwidth]{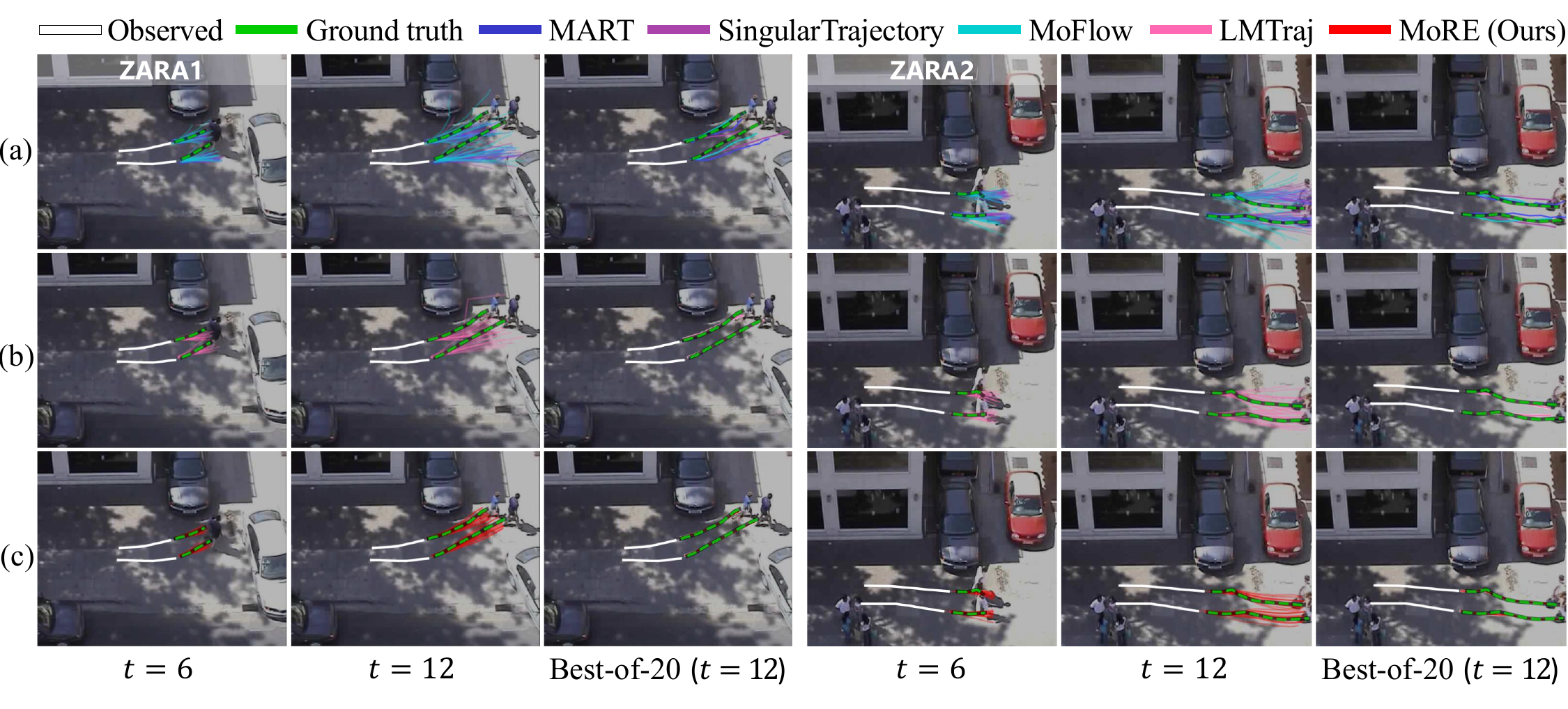}
\vspace{-8mm}
\caption{Sampled predictions on ZARA1 and ZARA2 at $t=6$ and $t=12$, with the best-of-20 trajectory. Rows show (a) numerical baselines, (b) LMTraj, and (c) MoRE. Additional qualitative results are provided in \cref{sec:supp_qualitative}.}
\label{fig:qualitative}

\vspace{2mm}

\includegraphics[trim={0 90mm 138 0}, clip, width=\linewidth]{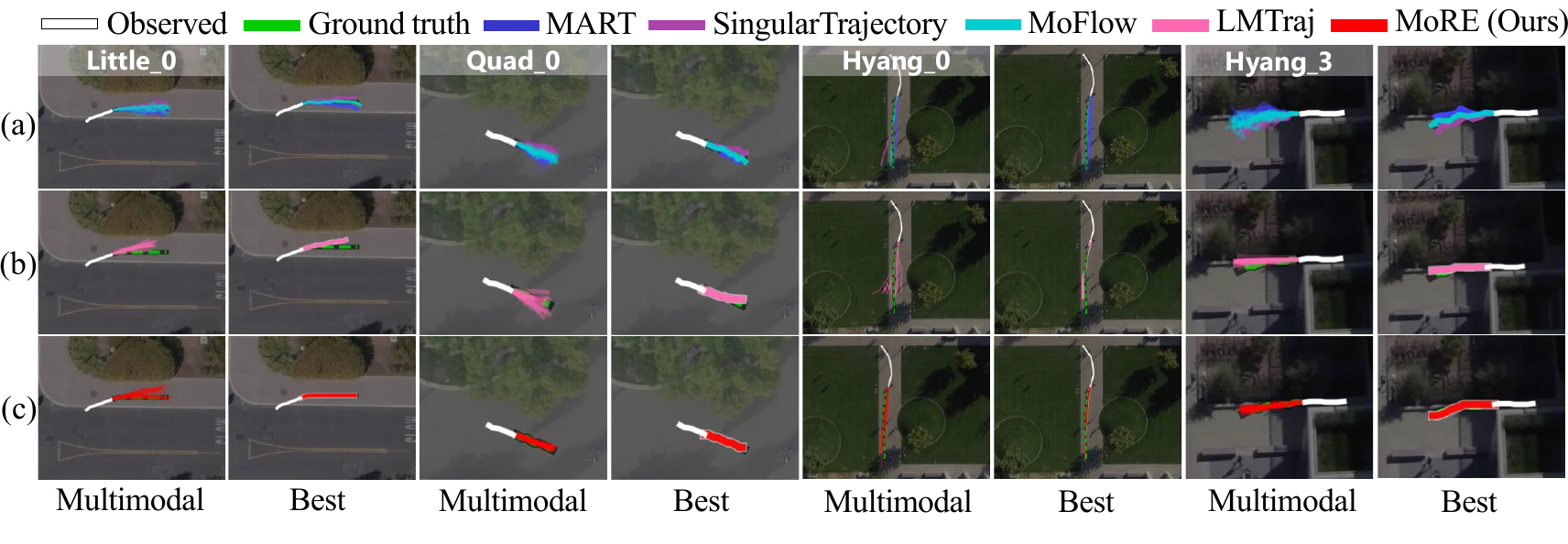}
\vspace{-8mm}
\caption{Sampled predictions on four SDD scenes, showing 20 trajectories and the best-of-20 trajectory. From top to bottom, rows show numerical baselines, LMTraj, and MoRE.}
\vspace{-4mm}
\label{fig:sdd_qualitative}

\end{figure}

\noindent\textbf{Interaction-heavy motion and social spacing.}
On the highly interactive NBA benchmark, MoRE achieves the lowest ADE and FDE among the compared methods (\cref{tab:nba_main}). Relative to LMTraj-SUP, ADE decreases from 1.10 to 0.96\,m and FDE from 1.21 to 1.11\,m, reductions of approximately 12.7\% and 8.3\%. On ETH-UCY, we further evaluate whether the predicted trajectories reproduce observed pedestrian spacing. MoRE is closest to the ground truth in all three distance zones (\cref{tab:combined}). The intimate-zone percentage decreases from 25.72\% to 23.41\%, closer to the ground-truth value of 22.48\%. COL also decreases from 2.54\% to 2.11\%, the lowest among the compared methods. Together, these results show that MoRE improves trajectory accuracy while producing sampled predictions with spacing patterns closer to real pedestrians and fewer collisions.

\noindent\textbf{Qualitative comparison.}
~\cref{fig:motivation}, \cref{fig:qualitative}, and~\cref{fig:sdd_qualitative} compare sampled predictions across ETH-UCY and SDD. In the illustrated cases, MoRE reduces the spatial offset of the language-based predictions and concentrates samples around paths closer to the ground truth. These qualitative observations complement the displacement and pedestrian-spacing measurements. Additional visual comparisons on ETH-UCY, SDD, and GCS are provided in \cref{sec:supp_qualitative}.

\subsection{Ablation Studies}
\label{sec:ablation}
\noindent\textbf{Expert integration and reward design.}
~\cref{tab:ablation_reward} examines both how numerical expert knowledge is integrated and how the reward is constructed. Additional SFT on the same hard-sample subset does not improve the base policy, while distillation remains above MoRE. Within PPO, maximum and mean aggregation give 0.218 / 0.314\,m and 0.217 / 0.307\,m, respectively, whereas UWO reaches 0.204 / 0.285\,m. Removing the ground-truth reward increases the error to 0.211 / 0.297\,m. Together, these results show that the gain depends on both the integration strategy and the proposed reward design, rather than additional fine-tuning or direct expert transfer alone. Output ensembling, additional non-PPO integration strategies, and per-scene results are reported in \cref{tab:integration_full}.

\begin{table}[!htbp]
\centering

\begin{minipage}[t]{0.465\linewidth}
\vspace{0pt}
\caption{Expert integration and reward ablation on ETH-UCY (ADE / FDE, meters). GT denotes ground truth.}

\label{tab:ablation_reward}
\end{minipage}%
\hfill%
\begin{minipage}[t]{0.515\linewidth}
\vspace{0pt}
\caption{Refinement with individual experts on ETH-UCY behavioral subsets (ADE, meters). Subsets may overlap; All combines the five experts.}
\label{tab:ablation_experts}
\end{minipage}

\par\smallskip

\begin{minipage}[t]{0.46\linewidth}
\vspace{0pt}
\centering
\small
\renewcommand{\arraystretch}{1.08}
\setlength{\tabcolsep}{20.5pt}

\begin{tabular}{@{}lrr@{}}
\toprule
\textbf{Method} & \textbf{ADE} & \textbf{FDE} \\
\midrule
Base policy
& 0.224 & 0.318 \\

Additional SFT
& 0.227 & 0.321 \\

Distillation
& 0.218 & 0.313 \\

PPO + Max
& 0.218 & 0.314 \\

PPO + Mean
& 0.217 & 0.307 \\

MoRE w/o GT
& \tul{0.211} & \tul{0.297} \\
\midrule

\textbf{MoRE (Ours)}
& \tbf{0.204} & \tbf{0.285} \\

\bottomrule
\end{tabular}
\end{minipage}%
\hfill%
\begin{minipage}[t]{0.51\linewidth}
\vspace{0pt}
\centering
\small
\renewcommand{\arraystretch}{1.08}
\setlength{\tabcolsep}{5.7pt}

\begin{tabular}{@{}lrrrr@{}}
\toprule
\textbf{Expert}
& \textbf{Group}
& \textbf{Mimicry}
& \textbf{Non-lin.}
& \textbf{Coll.} \\
\midrule

None
& 0.206
& 0.164
& 0.322
& 0.200 \\

STGCNN
& 0.202
& 0.158
& 0.312
& 0.215 \\

DMRGCN
& 0.198
& 0.148
& 0.318
& 0.205 \\

GP-Graph
& \tul{0.192}
& 0.152
& 0.310
& \tul{0.192} \\

SingularTraj.
& 0.212
& \tul{0.145}
& \tul{0.292}
& 0.202 \\

ExpertTraj.
& 0.210
& 0.160
& 0.308
& 0.198 \\
\midrule

\textbf{All (Ours)}
& \tbf{0.188}
& \tbf{0.142}
& \tbf{0.282}
& \tbf{0.185} \\

\bottomrule
\end{tabular}
\end{minipage}
\vspace{-3mm}
\end{table}

\noindent\textbf{Individual expert contributions.}
~\cref{tab:ablation_experts} evaluates refinement with individual experts on four behavioral subsets following prior work. These subsets may overlap. GP-Graph performs best among individual experts on group walking and collision avoidance, while SingularTrajectory performs best on mimicry and non-linear motion. The effect of each expert varies by behavior. For example, DMRGCN improves mimicry but does not improve the collision subset over the base policy. Combining all five experts gives the lowest ADE in every reported subset, showing the benefit of their complementary strengths. Full ADE / FDE and aggregate results are provided in \cref{tab:ablation_experts_full}.
\begin{table}[!htbp]
\centering
\begin{minipage}[t]{0.41\linewidth}\vspace{0pt}
\caption{Refinement on ETH-UCY (meters). Data is the training fraction used. Lower ADE / FDE is better.}
\label{tab:ablation_efficiency}
\end{minipage}\hfill
\begin{minipage}[t]{0.57\linewidth}\vspace{0pt}
\caption{Training samples before refinement. Base-policy errors are in meters. Coll.-inv. denotes collision-involved samples.}
\label{tab:selection_diagnostics}
\end{minipage}
\par\smallskip
\begin{minipage}[t]{0.41\linewidth}
\centering\small
\renewcommand{\arraystretch}{1}
\setlength{\tabcolsep}{5.33pt}
\begin{tabular}{lrr}
\toprule
\textbf{Tuning} & \textbf{Data (\%)} & \textbf{ADE / FDE} \\
\midrule
Full FT & 1 & \tul{0.213}/0.297 \\
LoRA & 100 & 0.221/0.303 \\
LoRA & 0.1 & 0.223/\tul{0.288} \\
\rowcolor{gray!10}
\textbf{LoRA (Ours)} & \textbf{1} & \tbf{0.204}/\tbf{0.285} \\
\bottomrule
\end{tabular}
\end{minipage}\hfill
\begin{minipage}[t]{0.57\linewidth}
\centering\small
\renewcommand{\arraystretch}{1}
\setlength{\tabcolsep}{3.91pt}
\begin{tabular}{lrrr}
\toprule
\textbf{Partition} & \textbf{ADE/FDE} & \textbf{Non-linear (\%)} & \textbf{Coll.-inv. (\%)} \\
\midrule
Top 1\% & 0.40/0.60 & 82 & 25 \\
Random 1\% & 0.24/0.35 & 45 & 13 \\
Bottom 1\% & 0.05/0.08 & 2 & 1 \\
All & 0.22/0.32 & 41 & 9 \\
\bottomrule
\end{tabular}
\end{minipage}
\vspace{-3mm}
\end{table}

\noindent\textbf{Sample selection and refinement.}
\cref{tab:ablation_efficiency} compares tuning strategies and data fractions. On the same 1\% subset, LoRA gives 0.204 / 0.285\,m, compared with 0.213 / 0.297\,m for full fine-tuning. Using 0.1\% or 100\% of the data raises LoRA's ADE to 0.223\,m or 0.221\,m. The 1\% setting is best among those shown. \cref{tab:selection_diagnostics} examines samples before refinement. The top 1\% has base-policy ADE / FDE of 0.40 / 0.60\,m, compared with 0.22 / 0.32\,m overall. Non-linear and collision-involved samples account for 82\% and 25\%, versus 41\% and 9\% overall. These results show that entropy ranking selects samples where the base policy struggles more and where non-linear and collision-related interactions occur more frequently. We therefore focus refinement on the top-entropy samples instead of applying numerical expert feedback to the entire training set. The random and bottom-entropy rows are included only to characterize sample difficulty, not as refinement baselines.

\begin{table}[!htbp]
\vspace{-3mm}
\centering
\caption{Additional evaluation. Forecasting entries are ADE / FDE. Non-linear errors are in meters and ETH-UCY-to-SDD transfer errors are in pixels. Inference is measured on an RTX 4090. The reference is the base policy for non-linear subsets and LMTraj-SUP for transfer and cost.}
\label{tab:additional_main}
\smallskip\small
\setlength{\tabcolsep}{31.02pt}
\begin{tabular}{lcc}
\toprule
\textbf{Evaluation} & \textbf{Reference} & \textbf{MoRE (Ours)} \\
\midrule
Non-linear group walking & 0.297 / 0.323 & \tbf{0.211 / 0.274} \\
Non-linear mimicry walking & 0.195 / 0.277 & \tbf{0.151 / 0.217} \\
Non-linear collision avoidance & 0.251 / 0.288 & \tbf{0.198 / 0.264} \\
Non-linear subset average & 0.248 / 0.296 & \tbf{0.187 / 0.252} \\
\midrule
ETH-UCY $\rightarrow$ SDD & 10.03 / 17.93 & \tbf{9.14 / 15.32} \\
\midrule
Inference memory (MB) & 1,401 & 1,401 \\
Inference latency (ms) & 18.3 & 18.3 \\
\bottomrule
\end{tabular}
\end{table}

\noindent\textbf{Hard cases, transfer, and inference cost.}
~\cref{tab:additional_main} evaluates MoRE only on challenging non-linear motion and interaction cases, using behavioral subsets that are distinct from the entropy-ranked training subset. MoRE improves both metrics in all three subsets, reducing the reported average ADE by approximately 24.6\%. This focused evaluation shows that the gains remain substantial on challenging non-linear motion and interaction cases. In ETH-UCY-to-SDD transfer without fine-tuning, FDE decreases from 17.93 to 15.32 pixels, approximately 14.6\%, showing that the refinement also transfers beyond the source dataset. Reported RTX 4090 measurements remain 1,401\,MB and 18.3\,ms, confirming that the numerical experts introduce no measured inference-time overhead. ~\cref{sec:supp_additional_experiments} provides full per-scene results, expert configurations, collision ablations, coefficient sweeps, and deterministic comparisons.

\FloatBarrier
\section{Conclusion}\label{sec:conclusion}
MoRE refines a pretrained language-based trajectory predictor using cached coordinate-level rewards from complementary numerical experts, uncertainty-weighted consensus, and entropy-based sample selection. Across ETH-UCY, SDD, GCS, and NBA, MoRE improves stochastic forecasting accuracy, including non-linear subsets and cross-dataset transfer. On ETH-UCY, it also yields pedestrian spacing closer to the ground truth and fewer collisions. Because experts are used only during training, measured inference memory and latency remain unchanged, while ablations show that expert composition and reward aggregation both matter.

\noindent\textbf{Limitations.}
MoRE depends on numerical experts, which may share errors or provide inaccurate forecasts. Expert disagreement can reflect valid alternative futures, so consensus does not guarantee safe or calibrated predictions. Deterministic gains remain small because available deterministic experts provide limited improvement, motivating stronger deterministic forecasters. Future work will investigate stronger deterministic experts and adaptive expert weighting to provide more reliable guidance. We also plan to extend MoRE to broader domains and evaluate multimodal coverage under larger distribution shifts.

\label{LastMainPage}
\clearpage

\subsection*{AI use statement}
Generative AI tools were not used for any activities requiring disclosure, including synthetic data generation, theoretical model development, the formulation or proof of mathematical claims, research methodology design, method implementation, or result interpretation. The authors drafted all text and used generative AI tools solely to improve its readability. All edited text was reviewed by the authors, who take full responsibility for the final content of this work.

\section*{Ethics Statement}
This work uses only existing, publicly available datasets for trajectory prediction.
These datasets may have limited coverage of motion patterns and environments, and the proposed method may inherit biases present in the source data.
Predicted trajectories are uncertain estimates rather than guaranteed outcomes of individual behavior.
Deployment in safety-critical or privacy-sensitive settings would require further validation and appropriate safeguards, particularly where predictions could support the tracking or monitoring of individuals.
Any materials released with this work will comply with the applicable licenses and redistribution restrictions of the source datasets.

\section*{Reproducibility Statement}
To support reproducibility, \cref{sec:method} presents the model formulation, learning objectives, sample selection, and implementation settings. \cref{sec:supp_rl_details} provides further details on the RL formulation, value network, optimization, hyperparameters, and numerical handling. \cref{sec:setup} describes the datasets, evaluation protocols, and metrics, while \cref{sec:supp_additional_experiments} reports additional ablations, sensitivity analyses, transfer experiments, and computational costs. Additional qualitative results are provided in \cref{sec:supp_qualitative}. We intend to release our code and evaluation scripts upon acceptance, subject to internal approval.

\bibliography{references}
\bibliographystyle{iclr2027_conference}
\clearpage
\appendix
\input{appendix}
\end{document}

%% file: appendix.tex
\section{RL Formulation and Implementation}\label{sec:supp_rl_details}
This appendix expands the training procedure in \cref{sec:method}. Extended related work appears in \cref{sec:extended_related}. Additional experiments, interpretation, and qualitative results follow in \cref{sec:supp_additional_experiments,sec:supp_discussion,sec:supp_qualitative}. The learning architecture, numerical reward definitions, and hyperparameters are retained from the main formulation.

\subsection{MDP Formulation for Token Generation}\label{sec:supp_rl_mdp}
Trajectory generation is an episodic Markov decision process. At token step $t$, the state is $s_t=(\tau_{\mathrm{obs}},\tau_{<t})$ and the action is the next token $a_t=\tau_t\in\mathcal V$. Numerical feedback is evaluated after the entire token sequence is decoded into coordinates. Intermediate rewards are zero. The terminal reward is the bounded ground-truth and expert return used for training:
\begin{equation}
 r_t=\begin{cases}0,&t<L,\\R,&t=L.\end{cases}
 \label{eq:mdp_reward}
\end{equation}
With discount factor $\gamma=1$, every token receives the same terminal return. The raw score is defined in \cref{eq:total_reward}, and its clipping is described below. Expert predictions are functions of the observed trajectory, not of the generated token prefix. Consequently, caching these predictions once avoids repeated expert inference during policy refinement. Decoding and coordinate-error calculations still occur for each rollout.

\subsection{Value Network and Shared Advantage}\label{sec:supp_rl_value}
The critic is a two-layer MLP with a ReLU activation, attached to the mean-pooled encoder features of the observation prompt, $\mathbf h_{\mathrm{obs}}\in\mathbb R^{d_{\mathrm{model}}}$. It predicts the terminal return from the initial context rather than the return from every generated prefix. The shared advantage estimate is
\begin{equation}
 \hat A=R-V_\phi(\mathbf h_{\mathrm{obs}}).
 \label{eq:shared_advantage}
\end{equation}
This is a sequence-return baseline, not a learned token-state value function or a generalized-advantage estimator. The actor's token-level terms share this scalar, while the critic is trained by regression to the return. The distinction matters because terminal-only feedback does not by itself require a context-only critic. It is the chosen lightweight baseline in this implementation.

\subsection{Token-Level Policy Objective}\label{sec:supp_rl_ppo}
The actor uses the clipped token-level surrogate in \cref{eq:ppo}. The likelihood ratio is relative to the rollout behavior policy $\pi_{\mathrm{old}}$, whereas the KL penalty is relative to the frozen pretrained reference $\pi_{\mathrm{ref}}$. These policies have different roles. The same sequence-level advantage is used for every token, and the actor objective is averaged over the generated length. The supervised objective instead averages over the ground-truth sequence length.

The training schedule uses one gradient update per rollout, interleaved with supervised updates. Under a strictly on-policy single-step update, with no parameter change between sampling and loss evaluation, the importance ratios initially equal one and clipping is locally inactive. Thus, using a PPO surrogate should not be interpreted as independently demonstrating a benefit from clipping. The empirical comparisons evaluate the complete reward-guided refinement procedure, including its supervised and reference-policy regularization, rather than isolating every optimizer component.

\subsection{Hyperparameters and Numerical Handling}\label{sec:supp_rl_hyperparams}
One PPO update is interleaved with every ten supervised steps (\texttt{rl\_step\_frequency}=10). The weights are $\lambda_{\mathrm{RL}}=0.5$, $\lambda_{\mathrm{exp}}=0.5$, and $\lambda_{\mathrm{uwo}}=0.05$. PPO uses clipping threshold $\epsilon=0.2$, value-loss coefficient $c_v=0.5$, and reference-KL coefficient $\beta=0.1$. Low-rank adapters have rank 16. AdamW uses a learning rate of $10^{-4}$ and an effective batch size of 1,024. The distributed configuration described in the manuscript uses eight RTX PRO 6000 GPUs with 128 samples per GPU.

Rollouts use temperature 0.7, top-$k$ sampling with $k=50$, and a maximum length of 100 tokens. The composite reward $R$ in \cref{eq:total_reward} is computed in coordinate space and then clamped to $[-10,0]$ before PPO optimization, with $-10$ assigned to malformed decoded sequences. Expert and ground-truth errors are computed in the tokenizer's normalized coordinate frame, so the same clipping range is used across benchmarks. The advantage clamp value is 5. In the optimization equations, $R$ denotes this bounded return used for training. The reported settings should be distinguished from test-time decoding, for which deterministic and stochastic evaluations are separate protocols.

\subsection{Regularization and Scope}\label{sec:supp_rl_stability}
The frozen reference, supervised token likelihood, low-rank adaptation, and interleaved update schedule are intended to limit changes to the pretrained distribution. They do not guarantee preservation of all auxiliary-task capabilities or prevent every form of reward exploitation. Our evidence consists of forecasting accuracy, pedestrian-spacing statistics, collision rates, and the evaluated transfer setting. We do not measure an independent semantic-reasoning capability before and after refinement.

\subsection{Geometry of the Mean Expert Reward}\label{sec:reward_geometry}
The following identity is an algebraic interpretation of the existing MSE reward, not an additional learning component. Let $d=2T_{\mathrm{pred}}$, $\bar{\mathbf S}=K^{-1}\sum_k\mathbf S_k$, and $C=K^{-1}\sum_k\|\mathbf S_k-\bar{\mathbf S}\|_F^2/d$. Expanding the squares gives
\begin{align}
 \frac1K\sum_{k=1}^K\MSE(\hat{\mathbf S},\mathbf S_k)
 &=\MSE(\hat{\mathbf S},\bar{\mathbf S})+C,\label{eq:mean_identity}\\
 \mu&=-\MSE(\hat{\mathbf S},\bar{\mathbf S})-C.\label{eq:mean_reward_identity}
\end{align}
The cross term vanishes because $\sum_k(\mathbf S_k-\bar{\mathbf S})=0$. For a fixed observation and its cached experts, $C$ is independent of the generated trajectory. The unclipped mean expert score therefore induces the same candidate ordering as distance to the average expert trajectory. It does not preserve the full multimodal predictive distribution of each expert.

The UWO term is the standard deviation of the scalar rewards $R_k(\hat{\mathbf S})$, which generally depends on the candidate. It penalizes unequal expert assessments of that candidate, not the fixed pairwise dispersion of cached expert trajectories alone. This explains why the mean-aggregation control is important. The identity does not imply that distillation and policy optimization have identical gradients, optimization dynamics, or regularization. Reward and advantage clipping further qualify any such equivalence.

\section{Extended Related Work}\label{sec:extended_related}
\subsection{Numerical Trajectory Prediction} 
Pioneering works~\citep{helbing1995social,pellegrini2009you,mehram2009socialforcemodel,yamaguchi2011you} rely on physics-based formulations such as social force models that describe pedestrian dynamics through attractive and repulsive forces.
With the advent of deep learning, Social-LSTM~\citep{alahi2016social} introduces data-driven recurrent modeling with social pooling to capture interactions between pedestrians.
Subsequent methods extend this paradigm by employing conditional variational autoencoders~\citep{lee2017desire,bhattacharyya2020conditional,mangalam2020pecnet,sun2021pccsnet,wang2022stepwise,xu2022groupnet,xu2022socialvae}, generative adversarial networks~\citep{gupta2018social,sadeghian2019sophie}, and attention mechanisms~\citep{fernando2018soft,ivanovic2019trajectron,salzmann2020trajectron++,vemula2018social}.
Graph-based architectures such as GCNs~\citep{mohamed2020social,shi2021sgcn}, GATs~\citep{bae2022npsn,huang2019stgat,kosaraju2019social,liang2020simaug} and transformers~\citep{yuan2021agent,bae2022npsn} enable direct modeling of pairwise interactions through spatial-temporal graphs, with extensions incorporating velocity- and distance-aware affinity~\citep{bae2021dmrgcn}.
Beyond pairwise interactions, group-aware methods~\citep{bae2022gpgraph,xu2022groupnet} explicitly model intra-group coherence and inter-group avoidance.
To capture generalizable motion patterns, recent approaches explore basis trajectory decomposition~\citep{bae2023eigentrajectory} with diffusion priors~\citep{bae2024singulartrajectory} that learn low-rank representations across diverse scenarios.
Meanwhile, goal-conditioned approaches~\citep{mangalam2020pecnet,zhao2021experttraj,dendorfer2020goalgan} leverage predicted destination information to guide trajectory interpolation, and diffusion models~\citep{gu2022mid,mao2023leapfrog} further enable better stochastic prediction.
Recent models further improve relational modeling and efficient forecasting~\citep{xu2023eqmotion,lee2024mart,fu2025moflow,chen2025pchgcn,li2025vite}.
These coordinate-based methods model interactions, groups, scene context, and motion patterns. Their different architectures provide complementary inductive biases~\citep{kothari2021human}. Complementary priors include pattern-guided forecasting~\citep{navarro2022social}, probabilistic B\'ezier curves~\citep{hug2020bezier}, interpretable social trees~\citep{shi2022social}, and neural social physics~\citep{yue2022nsf}, alongside group~\citep{bisagno2018group} and goal-directed prediction~\citep{rehder2015goal}.

\subsection{Language-based Trajectory Prediction}
Recently, inspired by the success of Large Language Models (LLMs) in high-level reasoning tasks~\citep{brown2020gpt3,wei2022cot,wei2022emergent} and driven by scaling laws~\citep{kaplan2020scaling,hoffmann2022chinchilla}, there have been attempts to incorporate language priors into trajectory forecasting.
These approaches typically recast trajectory sequences as text-based prompts, allowing the model to interpret coordinates as discrete tokens and formulate prediction as a generative question-answering task~\citep{bae2024lmtrajectory,bae2025vlmtrajectory,kim2025guidecot}.
To learn contextual representations, LMTraj learns auxiliary tasks such as collision assessment and group-member estimation~\citep{bae2024lmtrajectory}. Its multimodal extension instead uses a modality connector to incorporate visual information~\citep{bae2025vlmtrajectory}.
Recent methodologies also explore task-specific enhancements, such as fine-tuning models for egocentric pedestrian reasoning~\citep{shenkut2025visual} and employing goal-oriented visual prompts with chain-of-thought reasoning~\citep{kim2025guidecot}.
Some frameworks leverage LLMs as a submodule to incorporate high-level motion cues for numerical forecasting~\citep{chib2025lgtraj}.
More recent approaches directly formulate trajectory prediction as autoregressive language generation~\citep{wang2026autotraces} or optimize a language-based trajectory policy with reinforcement learning~\citep{xu2026w2w}, with related extensions to autonomous driving~\citep{jiang2026cort}.
MoRE focuses on transferring the predictions of frozen numerical forecasters into a pretrained language-based policy as reward feedback.
For predictors trained only with token likelihood, the objective does not directly penalize distances in continuous coordinate space.
MoRE supplies this feedback without changing the inference architecture.

\subsection{Multi-Expert Learning and Reward-based Optimization} 
Integrating insights from diverse models is a well-established paradigm for enhancing representational capacity, spanning from knowledge distillation (KD)~\citep{hinton2015distilling,you2017multiple} to Mixture-of-Experts (MoE) architectures~\citep{jacobs1991adaptive,shazeer2017moe,fedus2022switch}.
These mechanisms combine experts at the level of features or outputs, whereas MoRE uses numerical predictions to score trajectories sampled from the language model.
Reward learning from human preferences~\citep{christiano2017rlhf} supplies feedback beyond supervised targets. Subsequent work applies this approach to language-model alignment~\citep{ouyang2022instructgpt,bai2022harmless}, often using PPO~\citep{schulman2017ppo}.
While conventional alignment relies on human annotations, recent strategies increasingly formulate programmatic reward functions to explicitly enforce domain-specific constraints.
A critical issue in this monolithic reward optimization process is reward overoptimization~\citep{gao2023scaling}, wherein a policy exploits flaws in the proxy reward at the expense of the underlying task objective.
To mitigate this, \citet{coste2024reward} showed that ensembling reward models offers useful disagreement signals~\citep{lakshminarayanan2017deepensembles}, using disagreement for conservative policy updates.

Our use of multiple experts concerns the reward construction, rather than inference-time expert routing.
Consensus rewards and uncertainty-driven mining guide this transfer without inference-time experts.

\section{Additional Experimental Results}\label{sec:supp_additional_experiments}
\subsection{Sources and Reporting Conventions}\label{sec:sota_sources}
The references for \cref{tab:sota_comparison} are Social GAN~\citep{gupta2018social}, Social-STGCNN~\citep{mohamed2020social}, PECNet~\citep{mangalam2020pecnet}, Trajectron++~\citep{salzmann2020trajectron++}, AgentFormer~\citep{yuan2021agent}, MID~\citep{gu2022mid}, EqMotion~\citep{xu2023eqmotion}, GP-Graph~\citep{bae2022gpgraph}, LED~\citep{mao2023leapfrog}, MART~\citep{lee2024mart}, SingularTrajectory~\citep{bae2024singulartrajectory}, MoFlow~\citep{fu2025moflow}, PCHGCN~\citep{chen2025pchgcn}, ViTE~\citep{li2025vite}, LMTraj-SUP~\citep{bae2024lmtrajectory}, GUIDE-CoT~\citep{kim2025guidecot}, and W2W~\citep{xu2026w2w}. The NBA comparison follows \citet{wong2025resonance}.

Benchmark and ablation results are reproduced at their displayed precision. The main benchmark table uses the reporting precision shown for each benchmark, while detailed ablations retain higher-precision values when available. For example, the LMTraj-SUP ETH result is 0.409 / 0.501 in the detailed integration experiment and is reported as 0.41 / 0.50 in \cref{tab:sota_comparison}. Likewise, the ETH-UCY average is reported as 0.22 / 0.32 in the main benchmark table and 0.224 / 0.318 in the controlled integration comparison. Percentage reductions in the main text are approximate calculations from the displayed benchmark values. Behavioral subsets may overlap, so the reported aggregate in \cref{tab:ablation_experts_full} is evaluated separately and is not the arithmetic mean of the four displayed subset columns.

\subsection{Numerical Expert Selection}\label{sec:supp_expert_selection}

\begin{table}[H]
\centering
\caption{Numerical expert configurations on ETH-UCY (ADE / FDE, meters). The proposed five-expert model is the reference for leave-one-out, replacement, and addition tests.}
\label{tab:supplementary_experts}
\smallskip
\small
\setlength{\tabcolsep}{3pt}
\begin{tabular}{lr@{}}
\toprule
\textbf{Configuration} & \textbf{ADE / FDE} \\
\midrule
\multicolumn{2}{l}{\textit{(a) Single-expert alignment}} \\
SingularTrajectory & 0.211 / 0.315 \\
MART & 0.207 / 0.314 \\
MoFlow & 0.206 / 0.307 \\
\midrule
\multicolumn{2}{l}{\textit{(b) Leave-one-out removal}} \\
MoRE without Social-STGCNN & 0.207 / 0.291 \\
MoRE without DMRGCN & 0.206 / 0.291 \\
MoRE without GP-Graph & 0.208 / 0.293 \\
MoRE without SingularTrajectory & 0.208 / 0.291 \\
MoRE without Expert-Trajectory & 0.206 / 0.298 \\
\midrule
\multicolumn{2}{l}{\textit{(c) Replace SingularTrajectory}} \\
SingularTrajectory $\rightarrow$ MART & 0.207 / 0.289 \\
SingularTrajectory $\rightarrow$ MoFlow & 0.206 / 0.288 \\
\midrule
\multicolumn{2}{l}{\textit{(d) Add a sixth expert}} \\
MoRE + LED & 0.206 / 0.290 \\
MoRE + MART & 0.205 / 0.288 \\
MoRE + MoFlow & 0.205 / 0.287 \\
\midrule
\multicolumn{2}{l}{\textit{(e) Ground-truth reward ablation}} \\
MoRE without ground-truth reward & 0.211 / 0.297 \\
\textbf{MoRE (five experts)} & \textbf{0.204 / 0.285} \\
\bottomrule
\end{tabular}
\end{table}

Table~\ref{tab:supplementary_experts}(a) examines alignment with additional individual forecasters. In panel (b), removing any of the five experts raises both reported aggregate errors relative to the full configuration. Panels (c) and (d) show that replacing SingularTrajectory or adding a sixth expert does not improve the evaluated five-expert result. These tests support the selected combination. They do not establish that it is optimal among all available forecasters. Increasing the number of experts also increases the amount of prediction caching, even though the experts remain unnecessary at inference.

Panel (e), also summarized in \cref{tab:ablation_reward}, removes the ground-truth reward and gives 0.211 / 0.297, compared with 0.204 / 0.285 for the full reward. This supports retaining the ground-truth anchor. It is not the converse ablation of removing all expert rewards: the two questions should not be conflated.

\begin{table}[H]
\centering
\caption{Refinement with individual experts on ETH-UCY behavioral subsets (ADE / FDE, meters). Behavioral subsets may overlap, and the reported aggregate is evaluated separately from the four displayed subsets.}
\label{tab:ablation_experts_full}
\smallskip
\footnotesize
\setlength{\tabcolsep}{1.5pt}
\begin{tabular}{@{}lrrrrr@{}}
\toprule
\textbf{Expert} & \textbf{Group} & \textbf{Mimicry} & \textbf{Non-linear} & \textbf{Collision} & \textbf{Reported avg.} \\
\midrule
None & 0.206/0.256 & 0.164/0.232 & 0.322/0.474 & 0.200/0.275 & 0.223/0.309 \\
STGCNN & 0.202/0.252 & 0.158/0.212 & 0.312/0.455 & 0.215/0.292 & 0.221/0.301 \\
DMRGCN & 0.198/0.248 & 0.148/\tul{0.195} & 0.318/0.462 & 0.205/0.282 & 0.224/0.305 \\
GP-Graph & \tul{0.192}/\tul{0.242} & 0.152/0.202 & 0.310/0.450 & \tul{0.192}/0.268 & \tul{0.216}/0.298 \\
SingularTraj. & 0.212/0.268 & \tul{0.145}/0.200 & \tul{0.292}/\tul{0.428} & 0.202/0.278 & 0.218/0.296 \\
ExpertTraj. & 0.210/0.266 & 0.160/0.216 & 0.308/0.448 & 0.198/\tul{0.260} & 0.220/\tul{0.290} \\
\midrule
\textbf{All (Ours)} & \tbf{0.188}/\tbf{0.238} & \tbf{0.142}/\tbf{0.192} & \tbf{0.282}/\tbf{0.412} & \tbf{0.185}/\tbf{0.255} & \tbf{0.208}/\tbf{0.282} \\
\bottomrule
\end{tabular}
\end{table}
Table~\ref{tab:ablation_experts_full} reports both ADE and FDE for all five individual experts and their combination. Since the behavioral subsets may overlap, the reported aggregate is not obtained by averaging the four subset columns. Individual experts show different strengths across behaviors, while the combined reward gives the lowest ADE and FDE in all four displayed subsets and the lowest reported aggregate error. These subset results are distinct from the per-scene integration comparison in \cref{tab:integration_full}.

\subsection{Collision-Avoiding Performance}\label{sec:supp_col}

\begin{table}[H]
\centering
\caption{Collision rates on ETH-UCY (\%). Lower is better. The baseline values also appear in \cref{tab:combined}.}
\label{tab:collision}
\smallskip
\small
\setlength{\tabcolsep}{3pt}
\begin{tabular}{lrlr@{}}
\toprule
\textbf{Baseline} & \textbf{COL} & \textbf{Integration} & \textbf{COL} \\
\midrule
GP-Graph & 2.44 & Ensemble & 3.67 \\
LED & 2.62 & Knowledge distillation & 2.92 \\
MART & 2.23 & Maximum aggregation & 2.79 \\
SingularTrajectory & 2.96 & Mean aggregation & 2.37 \\
LMTraj-SUP & 2.54 & \textbf{UWO (MoRE)} & \textbf{2.11} \\
\bottomrule
\end{tabular}
\end{table}

Collision rates are computed across predicted multimodal samples under the protocol of \citet{liu2021snce}. Table~\ref{tab:collision} compares output/target transfer with reward aggregation. UWO has the lowest reported collision rate, followed by mean aggregation among the integration strategies. Maximum aggregation improves over the displayed output-level alternatives but is worse than the base policy. Thus, reward-based training alone is not sufficient to ensure improvement. The aggregation rule matters in these comparisons. The distance-zone measurements in \cref{tab:combined} assess agreement with observed spacing, not complete social compliance.

\subsection{Per-Scene Expert Integration}\label{sec:supp_integration}

\begin{table}[H]
\centering
\caption{Per-scene expert integration on ETH-UCY (ADE / FDE, meters). Aggregate values are reproduced at their reported precision.}
\label{tab:integration_full}
\smallskip
\footnotesize
\setlength{\tabcolsep}{3pt}
\begin{tabular}{lrrrrrr@{}}
\toprule
\textbf{Method} & \textbf{ETH} & \textbf{HOTEL} & \textbf{UNIV} & \textbf{ZARA1} & \textbf{ZARA2} & \textbf{Reported avg.} \\
\midrule
Base policy & 0.409/0.501 & 0.120/0.159 & 0.218/0.344 & 0.199/0.318 & 0.175/0.272 & 0.224/0.318 \\
Supervised FT & 0.401/0.514 & 0.108/0.148 & 0.247/0.353 & 0.211/0.319 & 0.168/0.271 & 0.227/0.321 \\
Ensemble & 0.421/0.612 & 0.173/0.261 & 0.292/0.514 & 0.231/0.382 & 0.203/0.331 & 0.264/0.420 \\
Distillation & 0.382/0.492 & 0.122/0.155 & 0.211/0.338 & 0.196/0.301 & 0.177/0.277 & 0.218/0.313 \\
Reward-weighted reg. & 0.388/0.499 & 0.121/0.158 & 0.215/0.342 & 0.198/0.305 & 0.173/0.283 & 0.219/0.317 \\
Minimum-risk training & 0.412/0.515 & 0.135/0.172 & 0.230/0.355 & 0.210/0.315 & 0.188/0.298 & 0.235/0.331 \\
Expert-guided FT & 0.386/0.495 & 0.124/0.157 & 0.213/0.339 & 0.196/0.302 & 0.186/0.287 & 0.221/0.316 \\
Max aggregation & 0.377/0.503 & 0.116/0.152 & 0.223/0.340 & 0.192/0.296 & 0.182/0.279 & 0.218/0.314 \\
Mean aggregation & 0.393/0.482 & 0.111/0.141 & 0.218/0.345 & 0.190/0.298 & 0.173/0.268 & 0.217/0.307 \\
UWO (MoRE) & 0.358/0.424 & 0.111/0.140 & 0.206/0.323 & 0.180/0.275 & 0.166/0.263 & 0.204/0.285 \\
\bottomrule
\end{tabular}
\end{table}

\Cref{tab:integration_full} includes every evaluated integration strategy. It expands the aggregate comparisons in \cref{tab:ablation_reward}, including reward-weighted regression, minimum-risk training, and expert-guided fine-tuning without PPO. UWO is the only evaluated configuration that improves both ADE and FDE over the base policy in every displayed scene. Other approaches have mixed effects. For example, additional supervised fine-tuning improves HOTEL but worsens UNIV. Mean aggregation improves several scenes but slightly worsens UNIV FDE. The controlled subset comparison supports the complete MoRE procedure but does not replace a matched-compute or multi-seed study of every baseline.

\subsection{Reward-Coefficient Sensitivity}\label{sec:supp_sensitivity}

\begin{table}[H]
\centering
\caption{Reward-coefficient sensitivity on ETH-UCY (ADE / FDE, meters), with $\lambda_{\mathrm{RL}}=0.5$. Columns vary $\lambda_{\mathrm{uwo}}$. Zero is mean aggregation.}
\label{tab:sensitivity_reward}
\smallskip
\small
\setlength{\tabcolsep}{3pt}
\begin{tabular}{lrrrrr@{}}
\toprule
\textbf{$\lambda_{\mathrm{exp}}$} & \textbf{0} & \textbf{0.01} & \textbf{0.05} & \textbf{0.1} & \textbf{0.5} \\
\midrule
0.1 & 0.223/0.316 & 0.218/0.309 & 0.211/0.297 & 0.213/0.300 & 0.222/0.315 \\
0.3 & 0.222/0.312 & 0.212/0.298 & 0.208/0.289 & 0.214/0.292 & 0.217/0.306 \\
0.5 & 0.217/0.307 & 0.210/0.295 & \textbf{0.204/0.285} & 0.206/0.288 & 0.211/0.299 \\
0.7 & 0.219/0.310 & 0.217/0.302 & 0.207/0.289 & 0.211/0.306 & 0.219/0.316 \\
1.0 & 0.221/0.309 & 0.216/0.300 & 0.210/0.294 & 0.215/0.297 & 0.221/0.312 \\
\bottomrule
\end{tabular}
\end{table}

\begin{table}[H]
\centering
\caption{Sensitivity to $\lambda_{\mathrm{RL}}$ on ETH-UCY (ADE / FDE, meters), with $\lambda_{\mathrm{exp}}=0.5$ and $\lambda_{\mathrm{uwo}}=0.05$.}
\label{tab:sensitivity_rl}
\smallskip
\footnotesize
\setlength{\tabcolsep}{3pt}
\begin{tabular}{lrrrrrr@{}}
\toprule
\textbf{$\lambda_{\mathrm{RL}}$} & \textbf{0 (SFT)} & \textbf{0.1} & \textbf{0.3} & \textbf{0.5} & \textbf{0.7} & \textbf{1.0} \\
\midrule
ADE / FDE & 0.227/0.321 & 0.213/0.301 & 0.207/0.290 & \textbf{.204/.285} & .206/.289 & .213/.303 \\
\bottomrule
\end{tabular}
\end{table}

Table~\ref{tab:sensitivity_reward} fixes $\lambda_{\mathrm{RL}}=0.5$ and varies expert weight and score-disagreement penalty. Table~\ref{tab:sensitivity_rl} fixes $\lambda_{\mathrm{exp}}=0.5$ and $\lambda_{\mathrm{uwo}}=0.05$ and varies the RL weight. The $5\times5$ grid and six RL weights contain 30 distinct configurations because the selected setting occurs in both tables. All are reported to converge without divergence or collapse. In the local range $\lambda_{\mathrm{exp}}\in[0.3,0.7]$ and $\lambda_{\mathrm{uwo}}\in[0.01,0.1]$, ADE ranges from 0.204 to 0.217.

At every tested expert weight, $\lambda_{\mathrm{uwo}}=0.05$ improves over mean aggregation ($\lambda_{\mathrm{uwo}}=0$). Setting $\lambda_{\mathrm{RL}}=0$ gives additional supervised fine-tuning on the same hard subset and does not explain MoRE's improvement. These experiments use the interleaved update schedule and reference regularization. They support sensitivity claims only over the tested coefficient range, not a general statement that reward balance cannot destabilize training.

\subsection{Complete Refinement-Efficiency Study}\label{sec:supp_tuning}
Table~\ref{tab:efficiency_full} retains all six settings behind the selected comparisons in \cref{tab:ablation_efficiency}. The 0.1\%, 1\%, 10\%, 50\%, and 100\% rows vary the fraction used for LoRA refinement. Full fine-tuning uses the same 1\% subset as the selected LoRA setting. LoRA on 1\% gives the lowest reported ADE and FDE. Its advantage over the 0.1\% setting is larger for ADE than for FDE, whereas expanding the subset does not consistently improve either metric.

\begin{table}[H]
\centering
\caption{Complete tuning and data-fraction comparison on ETH-UCY (ADE / FDE, meters). Parameter fractions refer to the forecasting backbone and exclude the critic.}
\label{tab:efficiency_full}
\smallskip\small
\begin{tabular}{@{}lrrrr@{}}
\toprule
\textbf{Tuning} & \textbf{Parameters (\%)} & \textbf{Data (\%)} & \textbf{ADE} & \textbf{FDE}\\
\midrule
Full FT & 100 & 1 & \tul{0.213} & 0.297\\
LoRA & 0.3 & 100 & 0.221 & 0.303\\
LoRA & 0.3 & 50 & 0.220 & 0.312\\
LoRA & 0.3 & 10 & 0.219 & 0.299\\
LoRA & 0.3 & 0.1 & 0.223 & \tul{0.288}\\
\midrule
\textbf{LoRA (Ours)} & 0.3 & 1 & \tbf{0.204} & \tbf{0.285}\\
\bottomrule
\end{tabular}
\end{table}

\subsection{Properties of Mined Samples}\label{sec:supp_entropy}

\begin{table}[H]
\centering
\caption{Entropy-partitioned training samples on ETH-UCY. ADE / FDE is measured with the base policy. Constant-velocity error uses pixel coordinates. Behavioral labels follow \citet{bae2024lmtrajectory}.}
\label{tab:entropy_partition}
\smallskip
\small
\setlength{\tabcolsep}{3pt}
\begin{tabular}{lrrrr@{}}
\toprule
\textbf{Metric} & \textbf{Top 1\%} & \textbf{Random 1\%} & \textbf{Bottom 1\%} & \textbf{All} \\
\midrule
Mean entropy (nats) & 2.8 & 1.3 & 0.4 & 1.3 \\
Base-policy ADE / FDE (m) & 0.40/0.60 & 0.24/0.35 & 0.05/0.08 & 0.22/0.32 \\
Constant-velocity error (px) & 11.0 & 4.8 & 0.8 & 4.5 \\
Non-linear (\%) & 82 & 45 & 2 & 41 \\
Collision-involved (\%) & 25 & 13 & 1 & 9 \\
\bottomrule
\end{tabular}
\end{table}

The top-entropy subset has larger base-policy errors, a higher non-linear fraction, and more collision-involved interactions than the full training set, as shown in \cref{tab:entropy_partition}. Constant-velocity extrapolation, which is independent of learned models, gives the same ordering. Its pixel-coordinate errors should not be directly compared in magnitude with the meter-coordinate ADE / FDE entries.

Entropy is used as a relative ranking statistic. A transformation that preserves the sample ordering leaves selection unchanged. Uncertainty errors that change the ordering can affect it. The random and bottom-entropy subsets in this table characterize data difficulty. They are not post-training comparisons of MoRE fitted to those subsets, and therefore do not independently establish the optimality of entropy-based selection over random selection.

\subsection{Cross-Dataset Transfer}\label{sec:supp_crossdata}

\begin{table}[H]
\centering
\caption{Cross-dataset transfer from ETH-UCY to SDD without fine-tuning on SDD. ADE / FDE is measured in pixels.}
\label{tab:cross_dataset}
\smallskip
\small
\setlength{\tabcolsep}{3pt}
\begin{tabular}{lrr@{}}
\toprule
\textbf{Model} & \textbf{ADE} & \textbf{FDE} \\
\midrule
LMTraj-SUP & 10.03 & 17.93 \\
\textbf{MoRE} & \textbf{9.14} & \textbf{15.32} \\
\bottomrule
\end{tabular}
\end{table}

The transfer values in \cref{tab:additional_main} are reproduced in \cref{tab:cross_dataset}. In this evaluation, the model is trained on ETH-UCY and tested on SDD without fine-tuning. The improvement from 17.93 to 15.32 pixels corresponds to a 14.6\% reduction in FDE. This supports transfer of the refined supervised backbone in this particular source-to-target setting. It is distinct from LMTraj-ZERO, which tests a general-purpose foundation model without trajectory-specific supervised training. One transfer experiment does not establish robustness to all domain shifts.

\subsection{Deterministic Evaluation}\label{sec:supp_deterministic}

\begin{table}[H]
\centering
\caption{Deterministic and best-of-20 results on ETH-UCY (ADE / FDE, meters).}
\label{tab:deterministic}
\smallskip
\small
\setlength{\tabcolsep}{3pt}
\begin{tabular}{lrr@{}}
\toprule
\textbf{Model} & \textbf{Deterministic} & \textbf{Stochastic} \\
\midrule
SocialVAE & 0.54 / 1.12 & 0.21 / 0.33 \\
EigenTrajectory & 0.51 / 1.11 & 0.21 / 0.34 \\
LMTraj-SUP & 0.477 / 0.880 & 0.22 / 0.32 \\
MoRE & 0.476 / 0.877 & 0.20 / 0.29 \\
MoRE + deterministic experts & 0.477 / 0.878 & 0.20 / 0.29 \\
\bottomrule
\end{tabular}
\end{table}

Table~\ref{tab:deterministic} reports deterministic and stochastic comparisons. MoRE's deterministic difference from LMTraj-SUP is 0.001\,m ADE and 0.003\,m FDE. Adding deterministic experts does not improve those reported values. These are small point-estimate differences rather than evidence of a substantial deterministic gain. The stochastic improvement is the main empirical result. A single sampled output should not automatically be equated with deterministic decoding. The column labels follow the reported evaluation modes.

\subsection{Non-Linear Behavioral Subsets}\label{sec:supp_nonlinear}

\begin{table}[H]
\centering
\caption{Non-linear behavioral subsets of ETH-UCY (ADE / FDE, meters). These subsets differ from the four categories in \cref{tab:ablation_experts}.}
\label{tab:nonlinear_subset}
\smallskip
\small
\setlength{\tabcolsep}{3pt}
\begin{tabular}{lrrrr@{}}
\toprule
\textbf{Method} & \textbf{\shortstack{Non-linear\\group walking}} & \textbf{\shortstack{Non-linear\\mimicry walking}} & \textbf{\shortstack{Non-linear\\collision avoidance}} & \textbf{\shortstack{Subset\\average}} \\
\midrule
Base policy & 0.297/0.323 & 0.195/0.277 & 0.251/0.288 & 0.248/0.296 \\
MoRE & 0.211/0.274 & 0.151/0.217 & 0.198/0.264 & 0.187/0.252 \\
\bottomrule
\end{tabular}
\end{table}

The non-linear group-walking, mimicry, and collision-avoidance results are also summarized in \cref{tab:additional_main}. These subsets provide a more focused test than overall scene averages. MoRE improves both reported metrics in every category. The reported average ADE falls from 0.248 to 0.187\,m, a 24.6\% reduction. These additional non-linear subsets are not interchangeable with the broader Group, Mimicry, Non-linear, and Collision categories in \cref{tab:ablation_experts}.

\subsection{NBA and Computational Cost}\label{sec:supp_nba}
\begin{table}[H]
\centering
\caption{Computational cost on ETH-UCY. Inference memory and latency are measured on an RTX 4090. Training times follow the reported setups and are not hardware-matched.}
\label{tab:computational_cost}
\smallskip\small
\setlength{\tabcolsep}{5pt}
\begin{tabular}{@{}lrrr@{}}
\toprule
\textbf{Model} & \textbf{Memory (MB)} & \textbf{Training (h)} & \textbf{Inference (ms)} \\
\midrule
PECNet & 1,733 & 0.3 & 57.0 \\
MID & 2,929 & 6.9 & 35.0 \\
AgentFormer & 9,639 & 22.0 & 8.2 \\
SocialVAE & 1,762 & 2.1 & 73.0 \\
LMTraj-SUP & 1,401 & 3.8 & 18.3 \\
\midrule
\textbf{MoRE (Ours)} & 1,401 & 3.8 + 1 & 18.3 \\
\bottomrule
\end{tabular}
\end{table}
The cost entries for LMTraj-SUP and MoRE also appear in \cref{tab:additional_main}. The NBA comparison in \cref{tab:nba_main} extends evaluation to highly interactive motion and improves ADE from 1.10 to 0.96. Table~\ref{tab:computational_cost} separately compares inference memory, latency, and reported training durations on ETH-UCY. Memory refers to inference-time GPU memory, not the full training footprint of the policy, reference model, optimizer, critic, or expert precomputation. The training durations are not a hardware-matched speed ranking. The distributed refinement configuration is described in \cref{sec:supp_rl_hyperparams}.

\subsection{Inference on Additional Hardware}\label{sec:supp_efficiency}

\begin{table}[H]
\centering
\caption{Additional inference measurements in FP32 with PyTorch and without deployment-specific optimization. These are predictor latencies, not end-to-end navigation measurements.}
\label{tab:embedded}
\smallskip
\small
\setlength{\tabcolsep}{3pt}
\begin{tabular}{lrr@{}}
\toprule
\textbf{Platform} & \textbf{Latency (ms)} & \textbf{Rate (Hz)} \\
\midrule
NVIDIA DGX Spark & 55 & 18.2 \\
NVIDIA AGX Orin & 90 & 11.1 \\
\bottomrule
\end{tabular}
\end{table}

Table~\ref{tab:embedded} reports full-FP32 PyTorch inference without deployment-specific optimization. MoRE runs at 11.1\,Hz on AGX Orin and 18.2\,Hz on DGX Spark. These measurements characterize the predictor on the tested hardware. They do not include sensing, tracking, planning, control, or all multi-agent workloads of a deployed robot. No closed-loop navigation claim is inferred from these latency values.

\section{Discussion and Scope}\label{sec:supp_discussion}
\subsection{Numerical Knowledge and Contextual Prediction}\label{sec:a1}
Numerical models and language-based models both learn from trajectories and may both encode social context. The relevant distinction here is the form of supervision: the base policy uses a token-level objective, whereas the experts provide decoded coordinate-space feedback. MoRE uses their architectural diversity as a source of priors. It does not identify which model understands an agent's true intention, and it does not treat visible group-like motion as proof that a model recognizes a particular social relationship.

\subsection{Preservation of the Base Policy}\label{sec:a2}
The reference KL penalty, supervised loss, and restricted refinement subset constrain changes to the pretrained policy in different ways. The first penalizes distributional deviation, the second retains the ground-truth token objective on the selected data, and the third restricts which training contexts receive updates. These are design choices for stability, not guarantees of semantic fidelity. Because only the selected subset is used for refinement, behavior outside that subset must be assessed empirically. Good transfer on SDD is one such test, not an exhaustive guarantee.

\subsection{Difference from Inference-Time Ensembling}\label{sec:a3}
An output ensemble combines numerical predictions during forecasting. MoRE instead computes numerical predictions before policy refinement and then deploys only the refined language-based policy. Distillation is also a training-time transfer method, so the absence of inference-time experts is not unique to MoRE. The relevant empirical question is whether the reward-based training procedure improves over these alternatives under the evaluated controls. The mean-reward identity in \cref{sec:reward_geometry} cautions against attributing the entire effect to information unavailable in an averaged expert target.

\subsection{Remaining Evaluation Boundaries}\label{sec:scope}
Best-of-20 displacement metrics can improve without improving calibrated probability estimates or covering all plausible futures. Collision and distance-zone statistics add useful but partial information. Lower collision rate can also be influenced by sample diversity. Likewise, agreement between several experts does not imply correctness if their errors are correlated. The reported point estimates do not provide confidence intervals across random seeds. These boundaries motivate future evaluation of mode coverage, expert correlation, and statistically reliable gains, rather than broader claims of semantic understanding or deployment safety.

\section{Additional Qualitative Results}\label{sec:supp_qualitative}
We provide additional trajectory visualizations from ETH-UCY, SDD, and GCS. Numerical and language-based predictions can differ in both spread and spatial offset. MoRE shows improved alignment in selected examples, but these illustrations do not quantify uncertainty calibration, obstacle clearance, or an agent's latent intent. All claims about accuracy and social spacing rely on the corresponding quantitative results.

\begin{figure*}[t]
\centering
\includegraphics[trim={0, 320, 0, 0},clip,width=\textwidth]{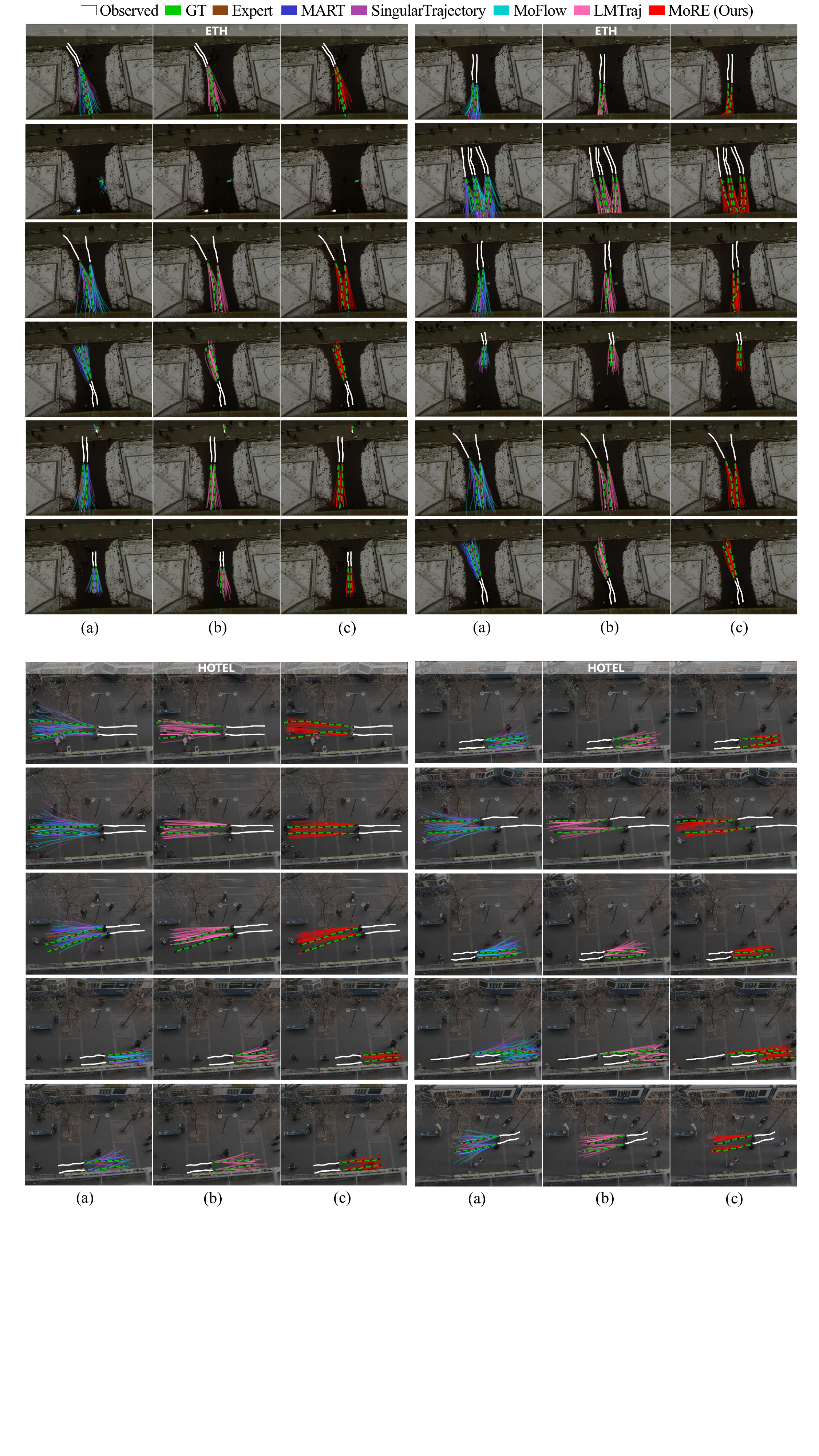}
\vspace{-7mm}
\caption{Visualization of prediction results on the ETH and HOTEL scenes. To aid visualization, selected agents are shown.}
\label{fig:supp_vis_1}
\end{figure*}

\begin{figure*}[t]
\centering
\includegraphics[trim={0, 280, 0, 0},clip,width=\textwidth]{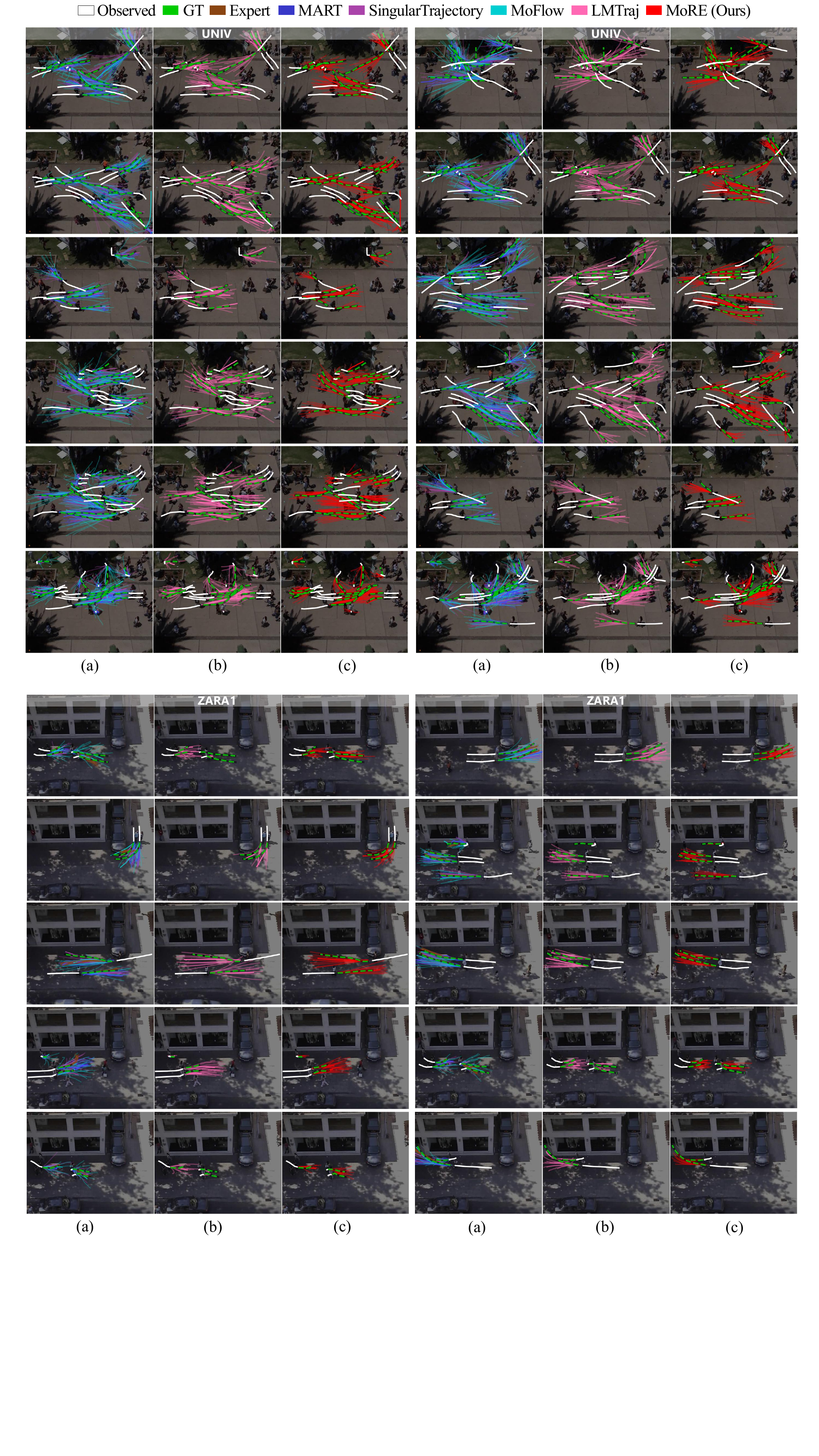}
\vspace{-7mm}
\caption{Visualization of prediction results on the UNIV and ZARA1 scenes. To aid visualization, selected agents are shown.}
\label{fig:supp_vis_2}
\end{figure*}

\begin{figure*}[t]
\centering
\includegraphics[trim={0, 1025, 0, 0},clip,width=\textwidth]{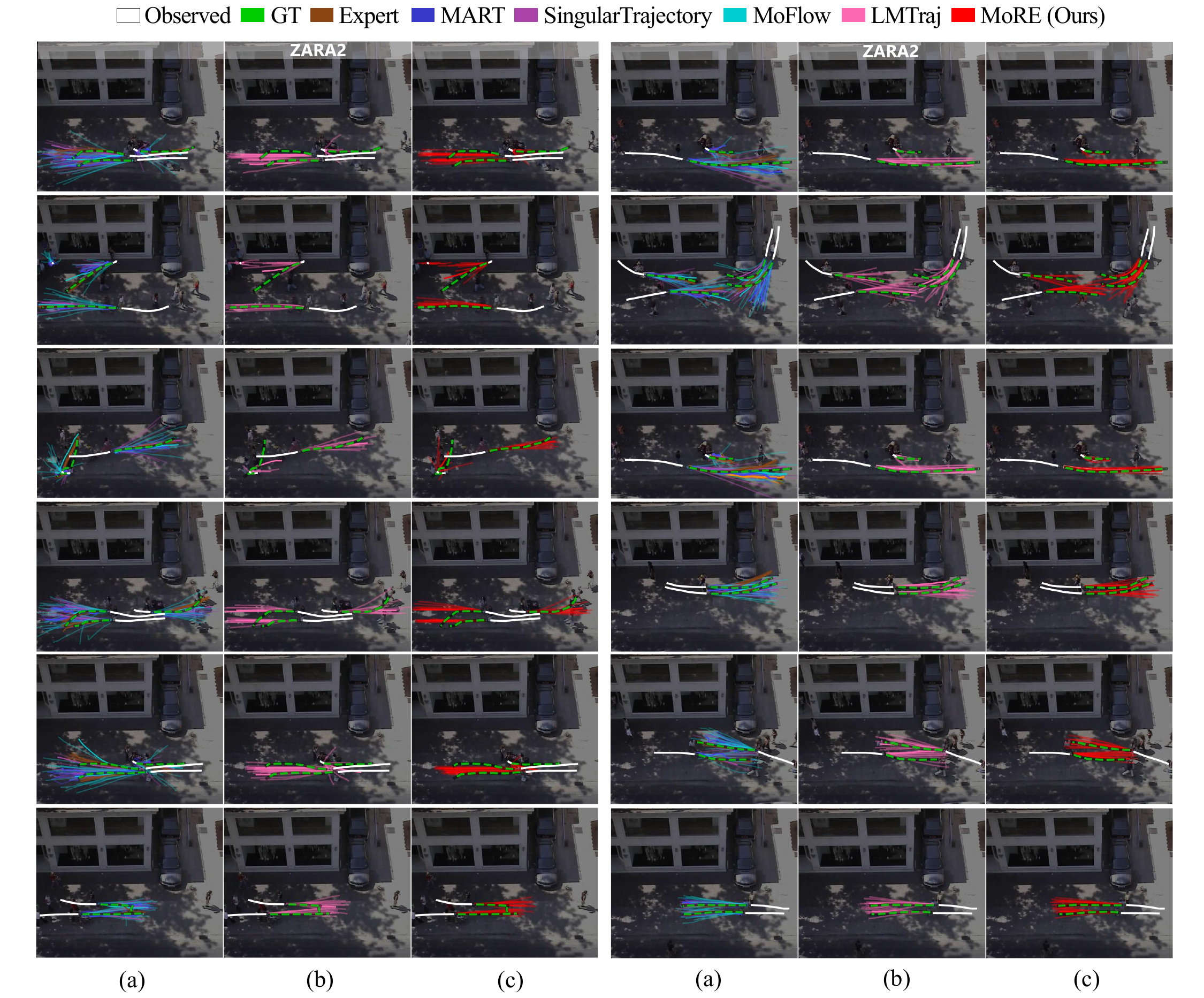}
\vspace{-6mm}
\caption{Visualization of prediction results on the ZARA2 scene. To aid visualization, selected agents are shown.}
\label{fig:supp_vis_3}
\end{figure*}

\begin{figure*}[t]
\centering
\includegraphics[trim={0, 1355, 0, 0},clip,width=\textwidth]{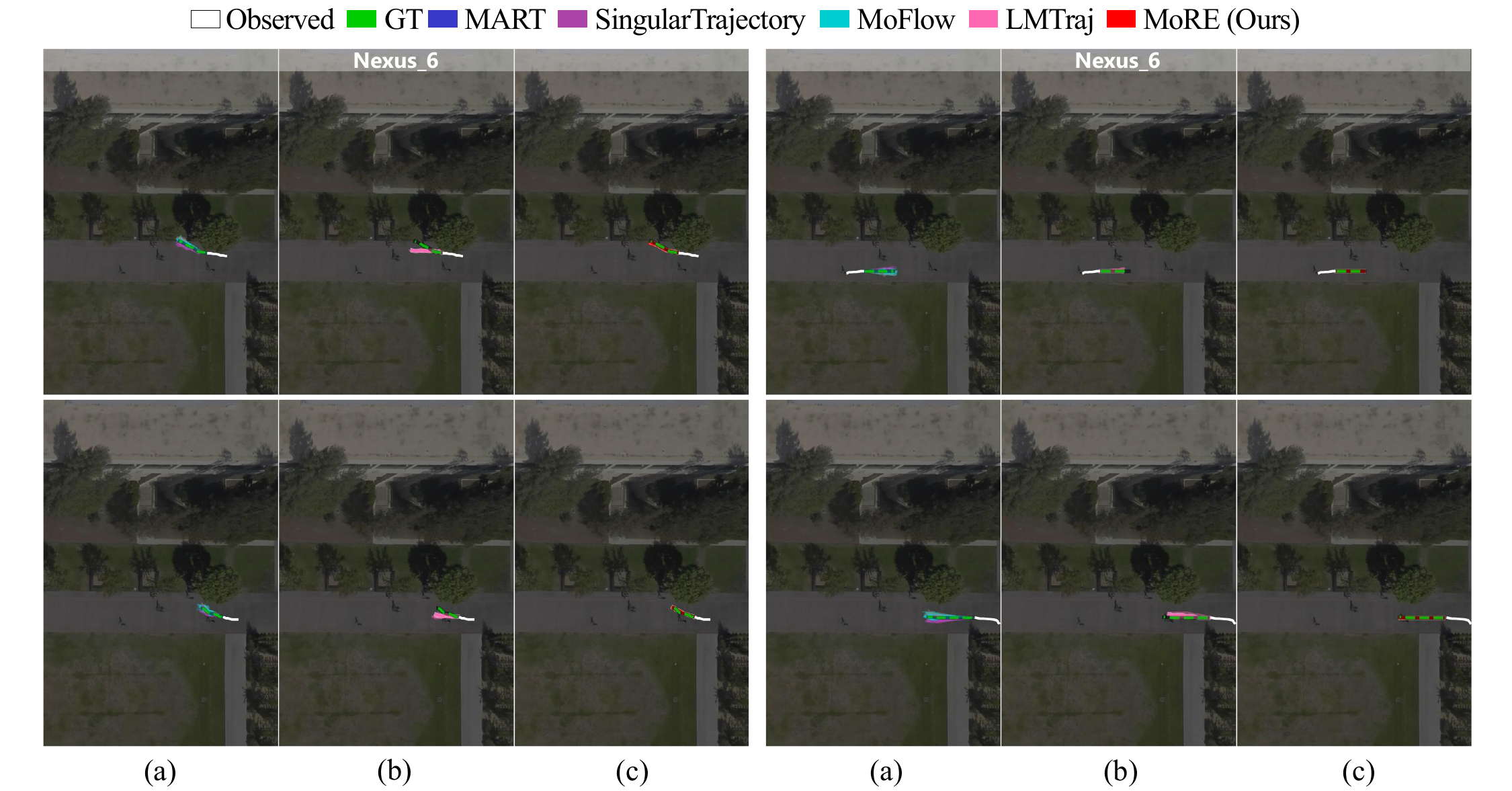}
\vspace{-6mm}
\caption{Visualization of trajectory predictions on nexus\_6 scene. To aid visualization, selected agents are shown.}
\label{fig:supp_vis_4}
\end{figure*}

\begin{figure*}[t]
\centering
\includegraphics[trim={50, 505, 50, 0},clip,width=\textwidth]{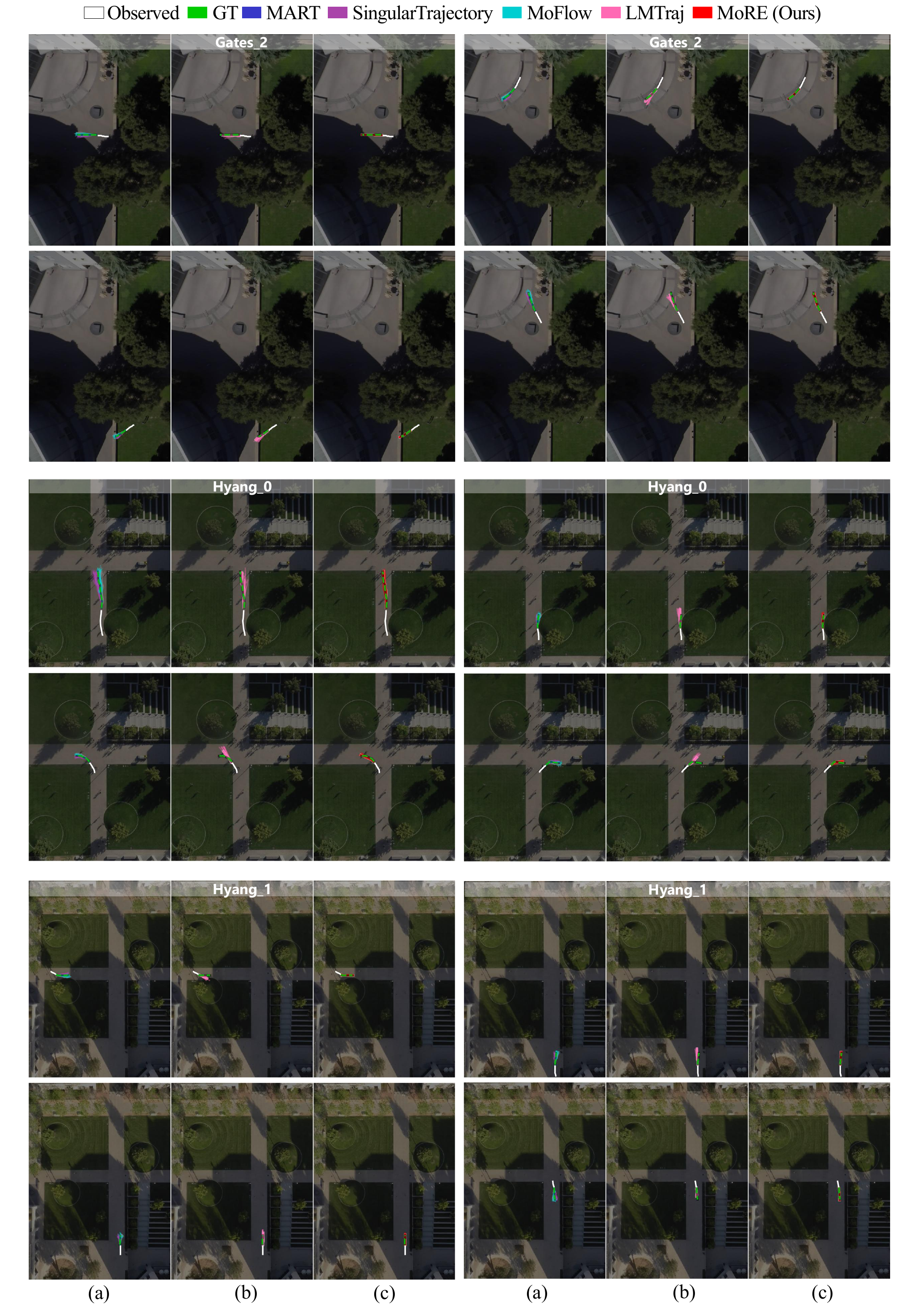}
\vspace{-8mm}
\caption{Visualization of trajectory predictions on gates\_2, hyang\_0, 1 scenes. To aid visualization, selected agents are shown.}
\label{fig:supp_vis_5}
\end{figure*}

\begin{figure*}[t]
\centering
\includegraphics[trim={25, 540, 25, 0},clip,width=\textwidth]{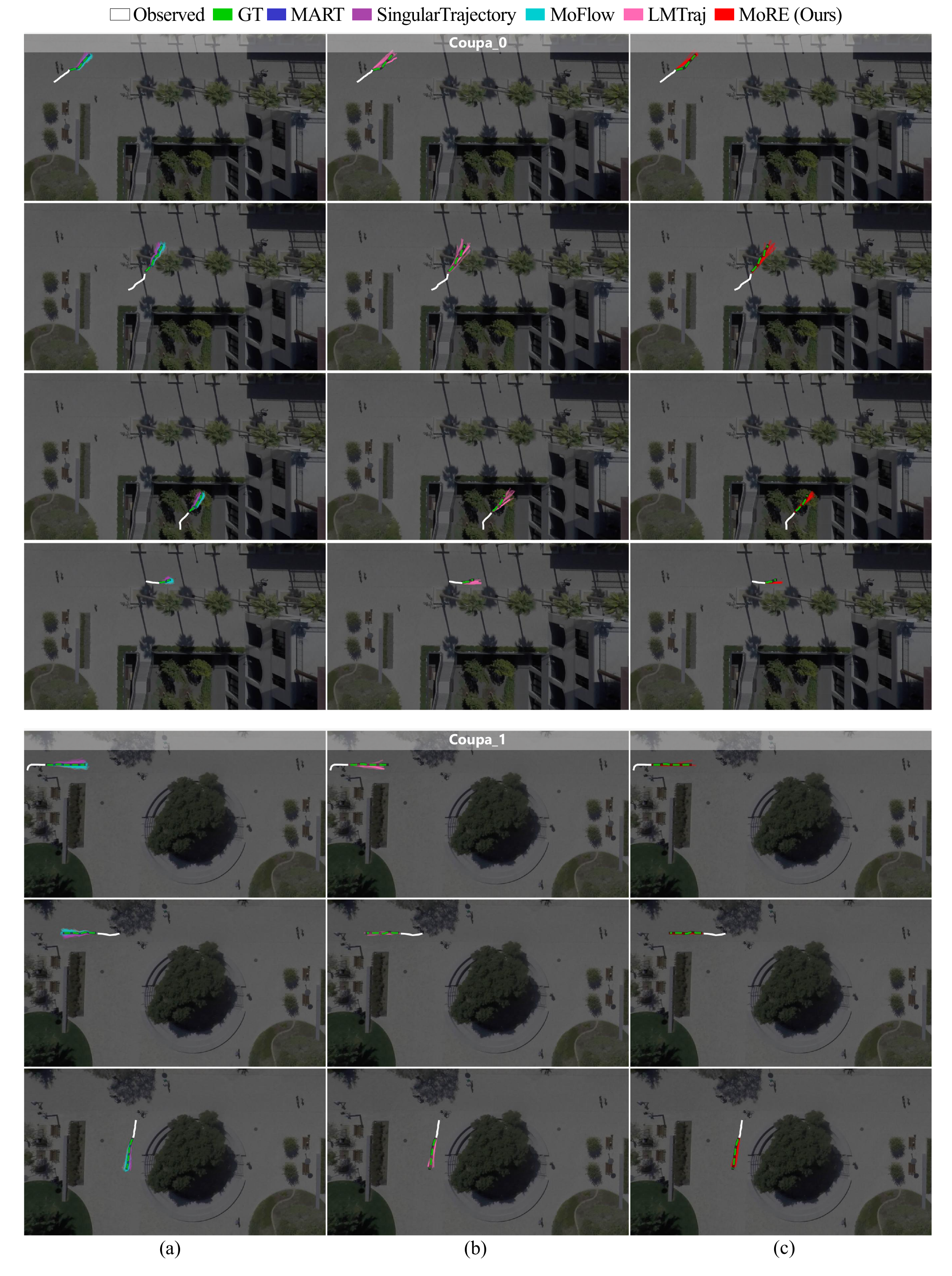}
\vspace{-9mm}
\caption{Visualization of prediction results on the coupa\_0 and coupa\_1 scenes. To aid visualization, selected agents are shown.}
\label{fig:supp_vis_6}
\end{figure*}

\begin{figure*}[t]
\centering
\includegraphics[trim={25, 575, 25, 0},clip,width=\textwidth]{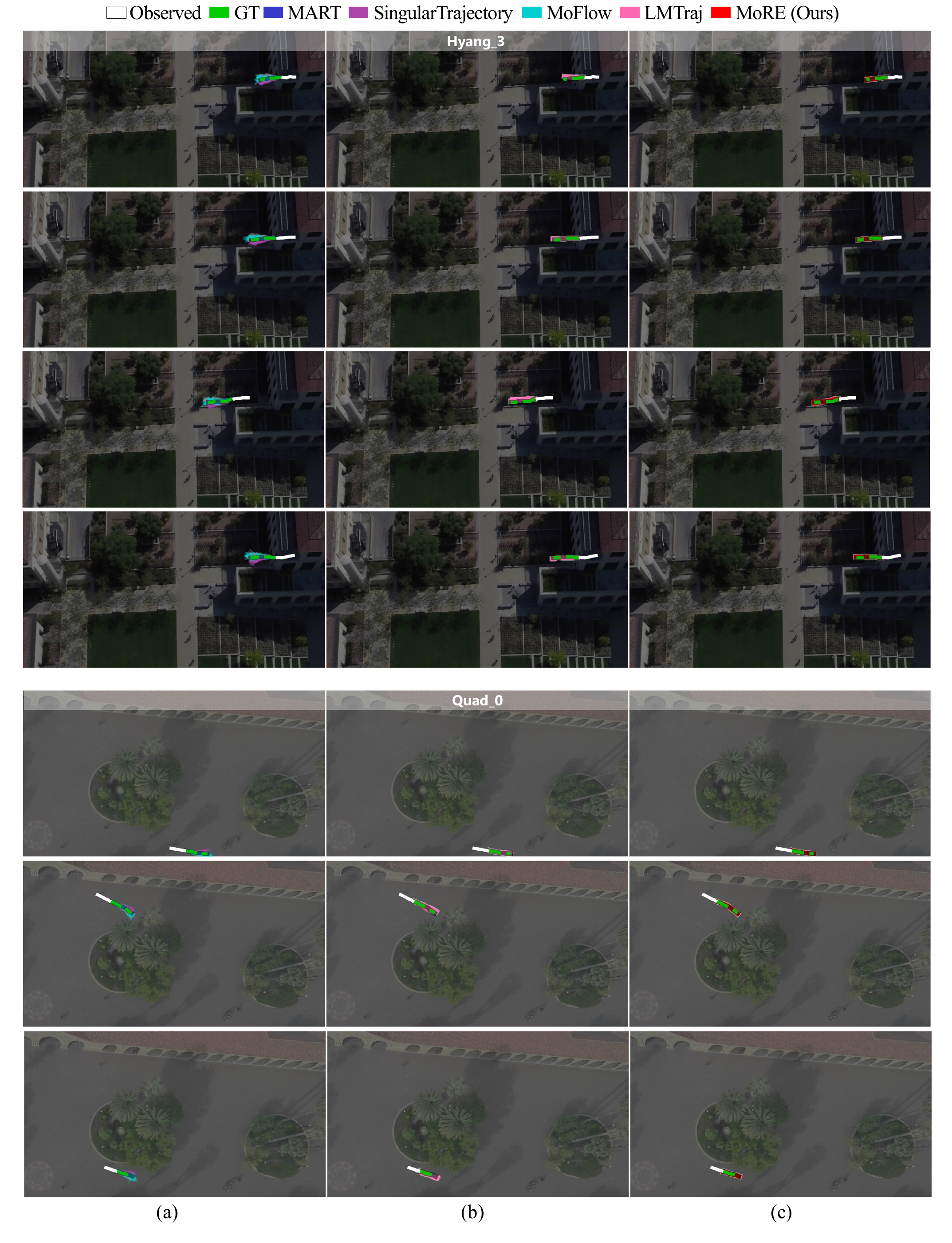}
\vspace{-9mm}
\caption{Visualization of trajectory predictions on hyang\_3 and quad\_0 scenes. To aid visualization, selected agents are shown.}
\label{fig:supp_vis_7}
\end{figure*}

\begin{figure*}[t]
\centering
\includegraphics[trim={25, 700, 25, 0},clip,width=\textwidth]{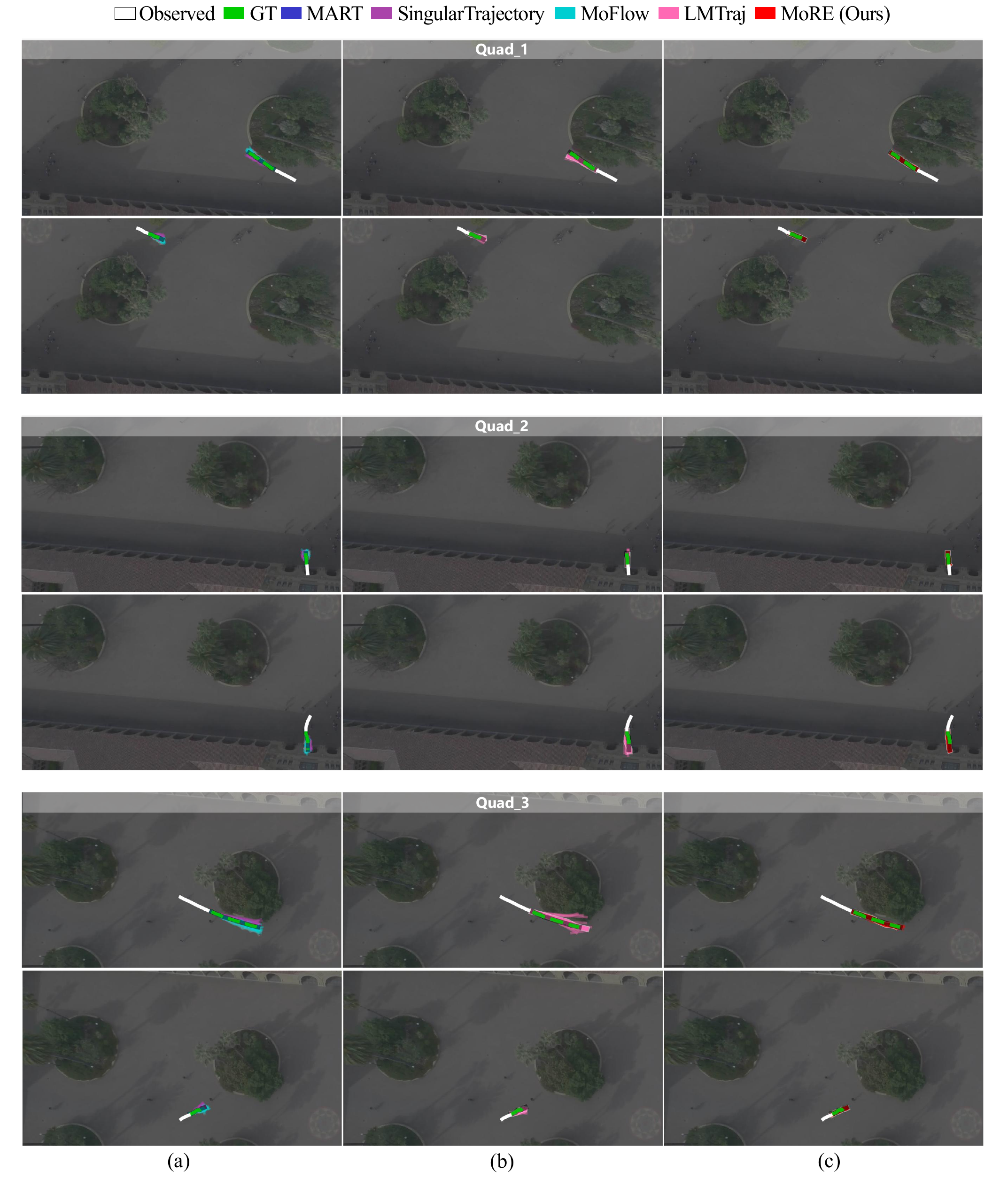}
\vspace{-9mm}
\caption{Visualization of prediction results on the quad\_1, 2, 3 scenes. To aid visualization, selected agents are shown.}
\label{fig:supp_vis_8}
\end{figure*}

\begin{figure*}[t]
\centering
\vspace{-5mm}
\includegraphics[trim={0, 320, 0, 0},clip,width=\textwidth]{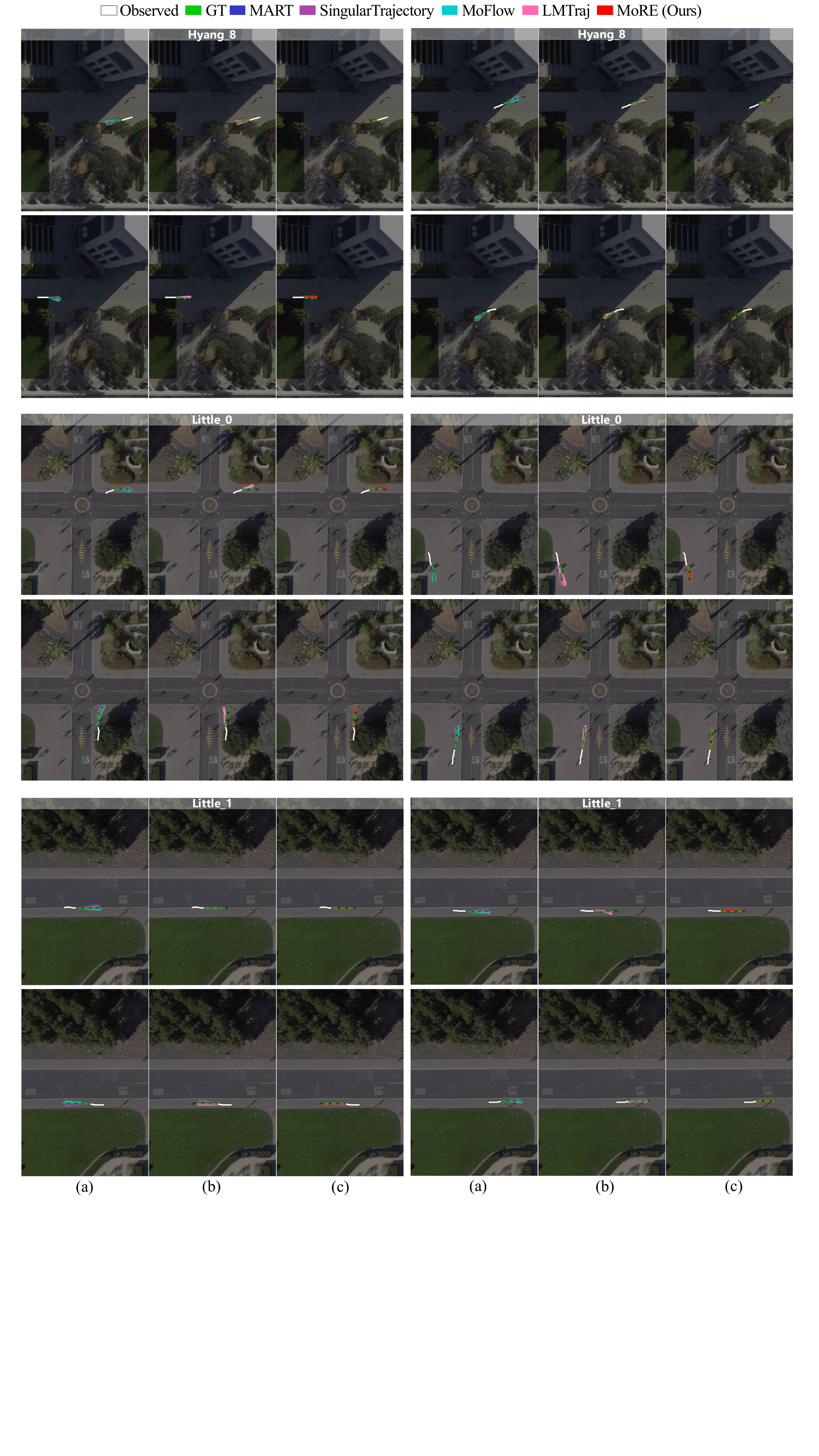}
\vspace{-9mm}
\caption{Visualization of prediction results on the hyang\_8 and little\_0, 1 scenes. To aid visualization, selected agents are shown.}
\vspace{-6mm}
\label{fig:supp_vis_9}
\end{figure*}

\begin{figure*}[t]
\centering
\vspace{-5mm}
\includegraphics[trim={0, 350, 0, 0},clip,width=\textwidth]{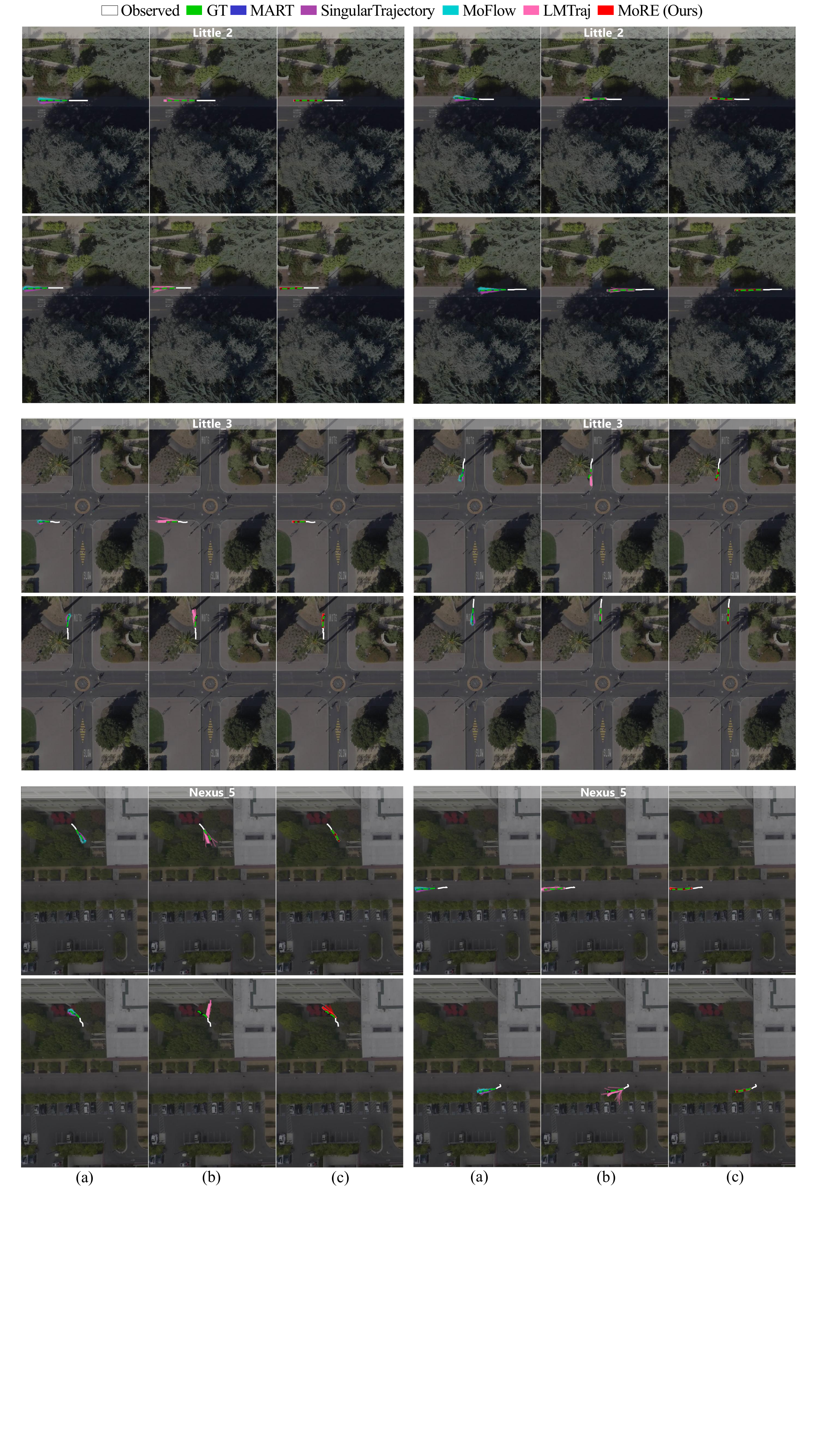}
\vspace{-7mm}
\caption{Visualization of prediction results on the little\_2, 3 and nexus\_5 scenes. To aid visualization, selected agents are shown.}
\vspace{-4mm}
\label{fig:supp_vis_10}
\end{figure*}

\begin{figure*}[t]
\centering
\includegraphics[trim={20, 510, 20, 0},clip,width=\textwidth]{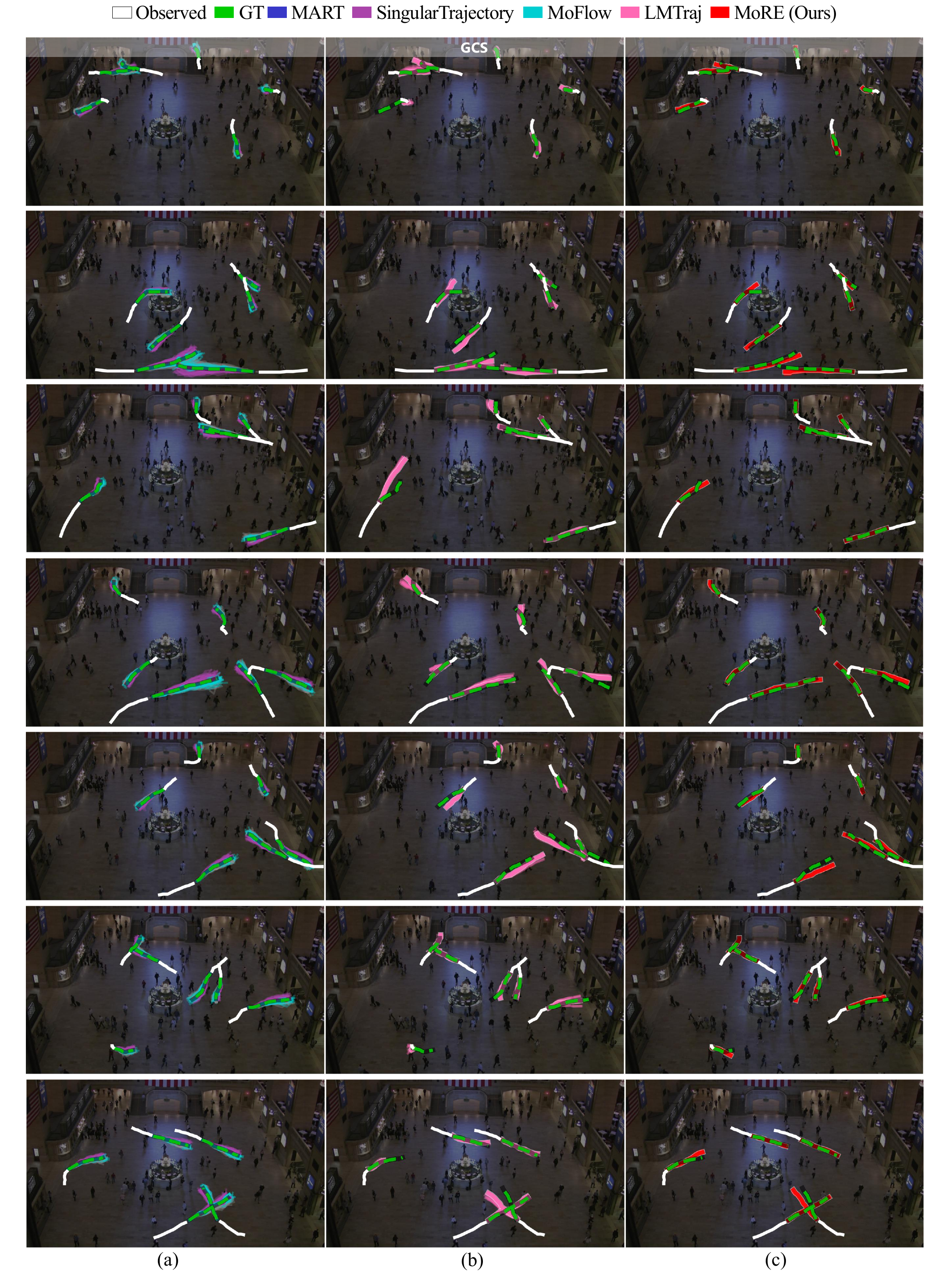}
\vspace{-9mm}
\caption{Visualization of prediction results on the GCS dataset. To aid visualization, selected agents are shown.}
\label{fig:supp_vis_11}
\end{figure*}

\clearpage

\clearpage